\documentclass{article}

\PassOptionsToPackage{numbers, compress}{natbib}

\usepackage[preprint]{neurips_2026}

\usepackage[utf8]{inputenc}
\usepackage[T1]{fontenc}
\usepackage{url}
\usepackage{booktabs}
\usepackage{amsfonts}
\usepackage{amsmath}
\usepackage{amssymb}
\usepackage{nicefrac}
\usepackage{microtype}
\usepackage{xcolor}
\usepackage{graphicx}
\usepackage{pifont}%
\definecolor{citecolor}{HTML}{0071BC}
\definecolor{linkcolor}{HTML}{ED1C24}
\definecolor{linkpink}{RGB}{210, 80, 140}
\usepackage[pagebackref=true,breaklinks=true,colorlinks,bookmarks=false,citecolor=citecolor,linkcolor=linkcolor,urlcolor=gray]{hyperref}

\usepackage{tikz}
\usepackage{pgfplots}
\pgfplotsset{compat=1.18}
\usepgfplotslibrary{groupplots}
\usetikzlibrary{
  arrows.meta,
  positioning,
  shapes.geometric,
  fit,
  calc,
  backgrounds,
  decorations.pathreplacing,
  patterns
}
\usepackage{subcaption}
\usepackage{multirow}
\usepackage{enumitem}
\usepackage{wrapfig}
\usepackage{algorithm}
\usepackage{algpseudocode}

\newcommand{\GraphDETR}{GraphDETR}
\newcommand{\BM}[1]{\boldsymbol{#1}}

\definecolor{tpgreen}{RGB}{34,139,34}
\definecolor{fpred}{RGB}{180,0,0}
\definecolor{blockblue}{RGB}{70,130,180}
\definecolor{blockorange}{RGB}{210,105,30}
\definecolor{blockgreen}{RGB}{60,160,60}

\newcommand{\ie}{\textit{i.e.,\ }}
\newcommand{\eg}{\textit{e.g.,\ }}

\newcommand{\cmark}{\ding{51}}%
\newcommand{\xmark}{\ding{55}}%

\title{End-to-End Subgraph Detection with \GraphDETR}

\author{%
  Dexiong Chen\textsuperscript{*}\qquad Till Hendrik Schulz\textsuperscript{*}\qquad Karsten Borgwardt\\[0.5 em]
  Max Planck Institute of Biochemistry, Martinsried, Germany\\
  \textsuperscript{*}Equal contribution\\[0.3em]
  {\hypersetup{urlcolor=linkpink}
  Code: will be released upon publication.}
}

\begin{document}

\maketitle

\begin{abstract}
\emph{Subgraph detection} seeks to identify whether and where instances of query patterns occur within a larger graph. This problem is fundamental across scientific domains and is closely related to subgraph isomorphism, which is NP-complete, limiting combinatorial approaches to small patterns or moderately sized graphs.
We introduce \GraphDETR{}, a deep learning framework that formulates subgraph detection as a set prediction problem, analogous to DETR in object detection.
\GraphDETR{} encodes the target graph with a graph neural network, and employs a fixed set of learnable query vectors, decoded via a transformer decoder, to predict all pattern occurrences jointly in a single forward pass.
This is enabled by training the model end-to-end with bipartite matching.
Unlike traditional combinatorial methods that only solve exact structural matching, \GraphDETR{} naturally extends to approximate matching, enabling detection beyond exact pattern correspondence.
Empirically, we show that \GraphDETR{} can detect diverse patterns, such as molecular structures, cycles, cliques, and fuzzy patterns of up to 50 nodes, in target graphs with up to 1000 nodes. 
We further evaluate on molecular functional group detection over the ChEMBL dataset, where \GraphDETR{} predicts the complete set of functional groups per molecule, achieving a strong performance of $\text{AP}_{100} = 91.2$.
\end{abstract}

\section{Introduction}
\label{sec:intro}

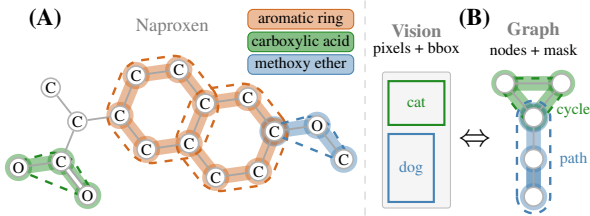
\begin{wrapfigure}{r}{7.7cm}
  \centering
  \vspace{-1em}
  \begin{tikzpicture}[
  atom/.style={circle, draw=gray!60, fill=white, minimum size=0.25cm, font=\tiny,
               inner sep=0pt, line width=0.6pt},
  panel/.style={font=\small\bfseries},
  lbl/.style={font=\scriptsize},
  leg/.style={draw, rounded corners=2pt, minimum width=1.40cm,
              align=center, font=\tiny, inner sep=1pt}
]

\begin{scope}[xshift=0cm]
  \node[panel]     at (0.42, 3.0) {(A)};
  \node[lbl, gray] at (2.10, 2.9) {Naproxen};

  \begin{scope}[yshift=0.3cm]

    \foreach \i/\cx/\cy in {
       5/1.408/1.425,  6/1.621/1.954,  7/2.186/2.033,
       8/2.537/1.584,  9/3.101/1.664, 10/3.452/1.215,
      13/3.239/0.686, 14/2.675/0.606, 15/2.323/1.056, 16/1.759/0.976}{
      \node[atom] (na\i) at (\cx,\cy) {C};
    }
    \node[atom] (na0)  at (0.492, 1.794) {C};
    \node[atom] (na1)  at (0.844, 1.345) {C};
    \node[atom] (na2)  at (0.630, 0.817) {C};
    \node[atom] (na3)  at (0.981, 0.367) {O};
    \node[atom] (na4)  at (0.066, 0.737) {O};
    \node[atom] (na11) at (4.017, 1.295) {O};
    \node[atom] (na12) at (4.368, 0.846) {C};

    \begin{scope}[on background layer]

      \foreach \ia/\ib in {5/6, 6/7, 7/8, 8/15, 15/16, 16/5,
                           8/9, 9/10, 10/13, 13/14, 14/15}{
        \draw[blockorange, opacity=0.45, line width=5.5pt, line cap=round]
          (na\ia)--(na\ib);
      }
      \draw[tpgreen, opacity=0.45, line width=5.5pt, line cap=round]
        (na2)--(na3);
      \draw[tpgreen, opacity=0.45, line width=5.5pt, line cap=round]
        (na2)--(na4);
      \draw[blockblue, opacity=0.45, line width=5.5pt, line cap=round]
        (na10)--(na11);
      \draw[blockblue, opacity=0.45, line width=5.5pt, line cap=round]
        (na11)--(na12);

      \foreach \cx/\cy in {
        1.408/1.425, 1.621/1.954, 2.186/2.033,
        2.537/1.584, 3.101/1.664,
        3.239/0.686, 2.675/0.606, 2.323/1.056, 1.759/0.976}{
        \filldraw[blockorange, fill opacity=0.45, draw=none]
          (\cx,\cy) circle (0.18cm);
      }
      \filldraw[blockblue,   fill opacity=0.45, draw=none]
        (3.452, 1.215) circle (0.2cm);   %
      \filldraw[blockorange, fill opacity=0.45, draw=none]
        (3.452, 1.215) circle (0.18cm);   %
      \foreach \cx/\cy in {0.630/0.817, 0.981/0.367, 0.066/0.737}{
        \filldraw[tpgreen, fill opacity=0.45, draw=none]
          (\cx,\cy) circle (0.18cm);
      }
      \foreach \cx/\cy in {4.017/1.295, 4.368/0.846}{
        \filldraw[blockblue, fill opacity=0.45, draw=none]
          (\cx,\cy) circle (0.18cm);
      }

      \foreach \ia/\ib in {5/6, 6/7, 7/8, 8/15, 15/16, 16/5}{
        \draw[gray!70, line width=0.7pt] (na\ia)--(na\ib);
      }
      \foreach \ia/\ib in {8/9, 9/10, 10/13, 13/14, 14/15}{
        \draw[gray!70, line width=0.7pt] (na\ia)--(na\ib);
      }
      \draw[gray!65, line width=0.7pt] (na5)--(na1);
      \draw[gray!65, line width=0.7pt] (na1)--(na0);
      \draw[gray!65, line width=0.7pt] (na1)--(na2);
      \draw[gray!70, line width=0.7pt, double distance=1.2pt] (na2)--(na3);
      \draw[gray!70, line width=0.7pt]                        (na2)--(na4);
      \draw[gray!70, line width=0.7pt] (na10)--(na11);
      \draw[gray!70, line width=0.7pt] (na11)--(na12);

      \draw[blockorange, dashed, line width=0.9pt, rounded corners=0.18cm]
        (2.755,1.615) -- (2.269,2.237) -- (1.485,2.128) -- (1.190,1.394)
        -- (1.677,0.772) -- (2.459,0.882) -- cycle;
      \draw[blockorange, dashed, line width=0.9pt, rounded corners=0.18cm]
        (3.670,1.246) -- (3.183,1.868) -- (2.401,1.758) -- (2.105,1.025)
        -- (2.593,0.402) -- (3.375,0.512) -- cycle;
      \draw[tpgreen, dashed, line width=0.9pt, rounded corners=0.25cm]
        (1.132,0.269) -- (0.697,0.985) -- (-0.111,0.772) -- cycle;
      \draw[blockblue, dashed, line width=0.9pt, rounded corners=0.25cm]
        (4.084,1.462) -- (3.275,1.249) -- (4.519,0.748) -- cycle;

    \end{scope}

  \end{scope}%

  \node[leg, fill=blockorange!45, draw=blockorange] at (3.8, 3.00) {aromatic ring};
  \node[leg, fill=tpgreen!55,     draw=tpgreen]     at (3.8, 2.70) {carboxylic acid};
  \node[leg, fill=blockblue!45,   draw=blockblue]   at (3.8, 2.40) {methoxy ether};
\end{scope}

\draw[gray!40, dashed, thin] (4.65, 0.40) -- (4.65, 3.3);

\begin{scope}[xshift=4.88cm]
  \node[panel] at (1.22, 3.0) {(B)};

  \node[font=\scriptsize\bfseries, gray] at (0.45, 2.85) {Vision};
  \node[font=\tiny, align=center] at (0.45, 2.6) {pixels + bbox};
  \draw[gray!50, fill=gray!8, rounded corners=1pt]
    (0.0, 0.5) rectangle (0.9, 2.3);
  \draw[tpgreen,   thick] (0.08, 1.60) rectangle (0.82, 2.18);
  \draw[blockblue, thick] (0.08, 0.58) rectangle (0.68, 1.48);
  \node[font=\tiny, tpgreen]   at (0.45, 1.89) {cat};
  \node[font=\tiny, blockblue] at (0.38, 1.03) {dog};

  \node[font=\large] at (1.22, 1.40) {$\Leftrightarrow$};

  \node[font=\scriptsize\bfseries, gray] at (2.0, 2.85) {Graph};
  \node[font=\tiny, align=center]        at (2.0, 2.6)  {nodes + mask};

  \foreach \i/\cx/\cy in {0/1.62/2.15, 1/2.37/2.15, 2/2.0/1.70}{
    \node[circle, draw=gray!60, fill=white,
          minimum size=0.25cm, inner sep=0pt, line width=0.6pt] (g\i) at (\cx,\cy) {};
  }
  \foreach \i/\cx/\cy in {3/2.0/0.65, 4/2.0/1.15}{
    \node[circle, draw=gray!60, fill=white,
          minimum size=0.25cm, inner sep=0pt, line width=0.6pt] (g\i) at (\cx,\cy) {};
  }

  \begin{scope}[on background layer]
    \foreach \ia/\ib in {0/1, 1/2, 2/0}{
      \draw[tpgreen,    opacity=0.45, line width=5.5pt, line cap=round] (g\ia)--(g\ib);
    }
    \draw[blockblue, opacity=0.45, line width=5.5pt, line cap=round] (g2)--(g3);
    \draw[blockblue, opacity=0.45, line width=5.5pt, line cap=round] (g3)--(g4);
    \foreach \cx/\cy in {1.62/2.15, 2.37/2.15, 2.0/1.70}{
      \filldraw[tpgreen,   fill opacity=0.45, draw=none] (\cx,\cy) circle (0.18cm);
    }
    \foreach \cx/\cy in {2.0/0.65, 2.0/1.15}{
      \filldraw[blockblue, fill opacity=0.45, draw=none] (\cx,\cy) circle (0.18cm);
    }
    \filldraw[blockblue, fill opacity=0.45, draw=none] (2.0,1.70) circle (0.2cm);
    \foreach \ia/\ib in {0/1, 1/2, 2/0}{
      \draw[gray!70, line width=0.7pt] (g\ia)--(g\ib);
    }
    \draw[gray!70, line width=0.7pt] (g3)--(g4);
    \draw[gray!65, line width=0.7pt] (g2)--(g4);  %
    \draw[tpgreen, dashed, line width=0.9pt, rounded corners=0.25cm]
      (2.537,2.217) -- (1.452,2.217) -- (2.002,1.520) -- cycle;
    \draw[blockblue, dashed, line width=0.9pt, rounded corners=0.22cm]
      (1.78,0.43) rectangle (2.22,1.92);
  \end{scope}

  \node[font=\tiny, tpgreen]   at (2.53, 1.8) {cycle};
  \node[font=\tiny, blockblue] at (2.53, 1.15) {path};

\end{scope}

\end{tikzpicture}
  \vspace{-0.5cm}
  \caption{%
    \textbf{(A)} A molecular graph with three color-coded functional groups, each predicted as a class label and a binary node mask. \textbf{(B)} The analogy between object detection and subgraph detection, where bounding boxes over pixels correspond to node masks over graphs.
  }
  \label{fig:teaser}
  \vspace{-1em}
\end{wrapfigure}

Subgraph detection is a fundamental problem in graph analysis that aims to identify whether and where specific patterns occur within a larger graph. 
It has applications in many domains such as molecular analysis~\citep{raymond2002heuristics,yang2007path}, and network science~\citep{plotnick1997concept,gentner1983structure}. 
A prominent instance of this problem is molecular functional group detection, where the goal is to locate all occurrences of chemically characteristic substructures such as hydroxyl or carbonyl groups within a molecule. 
Identifying these substructures plays a key role in tasks such as reaction prediction, retrosynthesis planning, and molecular property modeling. 
Figure~\ref{fig:teaser}A illustrates this task on an example molecule.

Classical approaches to locating pattern occurrences in graphs rely on combinatorial search algorithms for subgraph isomorphism, which seek injective edge-preserving mappings from a pattern to a target graph.
Methods such as \citet{ullmann1976} and \citet{cordella2004sub} provide exact solutions but suffer from exponential worst-case complexity, which limits their applicability on large graphs. %
Recent neural approaches~\citep{ying2021neuralsubgraph} attempt to address this limitation by learning representations of graphs or subgraphs using graph neural networks~(GNNs). 
However, most existing methods decompose the task into local matching decisions or independent predictions of candidate substructures. 
As a result, they do not explicitly model the fact that a target graph typically contains a \emph{set of subgraphs} that should be predicted jointly while avoiding duplicate detections.

Subgraph detection thus calls for jointly predicting all occurrences in a target graph, each as a labeled set of nodes, without duplicate detections. 
Casting the problem this way reveals a direct parallel with object detection, where the goal is likewise to predict a variable-sized set of objects, each with a class and a location, without duplicates (Figure~\ref{fig:teaser}B). 
In fact, recent progress in computer vision has shown that object detection can be solved elegantly as set prediction. 
In particular, DETR~\citep{carion2020detr} demonstrated this by training a fixed pool of learnable queries via a transformer decoder with bipartite matching.
DETR removes many heuristic components of traditional detection pipelines and enables fully end-to-end training.

In this work, we bring the set prediction perspective to graphs and introduce \GraphDETR{}, an end-to-end framework for subgraph detection. 
\GraphDETR{} first encodes the target graph using a GNN. 
A fixed set of learnable query vectors then interacts with the graph representation through a transformer decoder to produce candidate subgraphs. 
Each query predicts a node-set mask representing a detected subgraph. 
\GraphDETR{} is trained end-to-end with a bipartite matching loss that assigns predictions to ground-truth subgraphs, producing the complete set of detections in a single forward pass without duplicate predictions.
Unlike combinatorial methods, which are inherently limited to exact structural matching, \GraphDETR{} naturally extends to approximate matching, where a query pattern defines a class of subgraphs rather than a single exact structure.

We evaluate \GraphDETR{} in two settings. 
First, we apply \GraphDETR{} to molecular functional group detection, where the task is to predict the complete set of functional groups present in a molecular graph. Experiments on the ChEMBL dataset show that \GraphDETR{} achieves strong performance while maintaining a simple, unified, and fully end-to-end architecture.
Second, we consider subgraph matching with a fixed query set, which provides a general benchmark for evaluating the model’s ability to predict multiple substructures simultaneously.
Overall, our approach demonstrates that the set prediction paradigm provides a natural and effective formulation for graph reasoning problems involving multiple structured outputs. By combining GNNs with transformer-based set prediction, \GraphDETR{} provides a new approach for learning to detect subgraphs directly from data.

\section{Related Work}
\label{sec:related}

\GraphDETR{} is related to three lines of machine learning research: set-prediction-based detection, graph representation learning, and subgraph matching.

\textbf{Set prediction. }
DETR~\citep{carion2020detr} reformulated object detection as direct set prediction, where a transformer decoder with a fixed pool of learnable query vectors attends to image features and is trained end-to-end via Hungarian bipartite matching~\citep{kuhn1955hungarian}, eliminating anchors and non-maximum suppression.
Subsequent work refined query initialization~\citep{zhang2022dino} and extended the paradigm to instance segmentation~\citep{cheng2022mask2former}.
SAM~\citep{kirillov2023sam} and SAM~2~\citep{ravi2024sam2} introduced a \emph{two-way} transformer decoder in which prompt tokens and image tokens mutually attend to each other, allowing the image representation to incorporate per-query context before mask logits are computed.
\GraphDETR{} applies this full pipeline to graphs, where image feature maps become per-node embeddings produced by a graph encoder, bounding boxes become binary node-membership masks, and class labels become subgraph types.
The principal graph-specific addition is the cut penalty, which steers each predicted mask toward a spatially connected subgraph, a structural prior with no direct image analogue.

\textbf{Graph neural networks. }
The dominant paradigm for graph learning is message passing, with representative MPNNs including GCN~\citep{kipf2017gcn}, GIN~\citep{xu2019gin}, and GatedGCN~\citep{bresson2017gatedgcn}. 
Graph Transformers~\citep{ying2021graphormers,chen2022sat} augment local message passing with global self-attention.
GraphGPS~\citep{rampasek2022gps} runs an MPNN branch and self-attention in parallel per layer, attaining strong results on several benchmarks.
Random Walk Neural Networks~\citep{ivanov2018anonymous,tonshoff2023crawl}, in particular NeuralWalker~\citep{chen2025neuralwalker}, take an alternative route by encoding random-walk sequences with a sequence model, such as a state-space model, and aggregating back to nodes, capturing long-range structural context without the quadratic cost of full attention.
All standard MPNNs and Graph Transformers suffer from the symmetry issue, i.e., they assign identical representations to nodes that are in the same orbit.
This positional ambiguity is especially problematic for instance-level tasks, where two nodes belonging to \emph{different} occurrences of the same subgraph query may receive identical node embeddings, obstructing per-instance mask assignment.

\textbf{Subgraph matching. }
Classical algorithms, such as Ullmann's backtracking~\citep{ullmann1976} and VF2~\citep{cordella2004sub}, compute exact subgraph isomorphisms but run in worst-case exponential time, making them impractical on large graphs. %
Early neural approaches sidestep combinatorial search by learning graph-level similarity scores. 
For instance, Graph Matching Networks~\citep{li2019gmn} employ cross-graph attention to compute pairwise graph similarity, and SimGNN~\citep{bai2019simgnn} combines graph-level embeddings with node-to-node interaction to approximate graph edit distance.
These methods measure how similar two graphs are globally, but do not localize where in a larger graph a pattern occurs.
Neural Subgraph Matching~\citep{ying2021neuralsubgraph} learns order-preserving embeddings that determine whether a query is a subgraph of a target and can identify a matching neighborhood through post-processing. However, it only processes a single query–target pair and does not predict multiple simultaneous occurrences.
A related line of work learns to \emph{count} pattern occurrences~\citep{pellizzoni2025count}, returning a scalar estimate of how many times a query appears, without spatial localization.
\emph{\GraphDETR{} is the first model to jointly produce a complete set of query-subgraph pairs for all instances in a single differentiable forward pass, with no combinatorial search or post-processing at inference time.}

\section{GraphDETR for End-to-End Subgraph Detection}
\label{sec:method}

The task of subgraph detection requires identifying each occurrence of each pattern separately. 
The same pattern may appear \emph{multiple} times in one graph, and different occurrences may have \emph{overlapping node sets}. 
For each occurrence, the model needs to predict which pattern class it belongs to and which nodes it occupies. 
A suitable model should therefore produce a variable-size set of (class, node-mask) pairs, one per instance.

This prediction problem has a natural analogue in object detection, where a model must similarly produce a variable-size set of (class, localization) pairs, one per detected object. 
The established paradigm for this task, pioneered by DETR \citep{carion2020detr}, is phrased as set prediction, where a fixed set of learnable queries each produce one candidate. 
\GraphDETR{} adopts this formulation for graphs, replacing bounding boxes with binary node-membership masks and class labels with pattern types.

\begin{figure*}[t]
  \centering
  \input{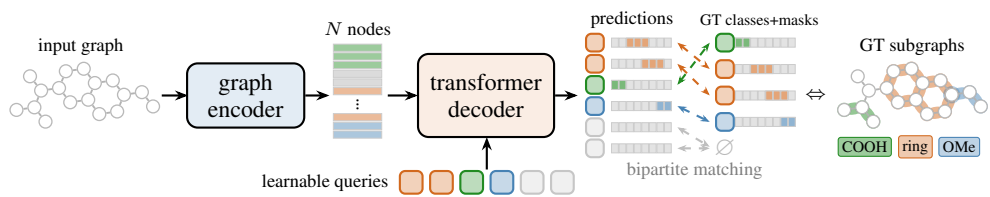}
  \caption{%
    \GraphDETR{} architecture.
    An input graph is encoded into per-node embeddings by the graph encoder.
    A set of learnable query vectors and the node embeddings are jointly refined by a (two-way) transformer decoder via interleaved bidirectional cross-attention.
    A linear class head and a bilinear mask head produce one (class, node-mask) pair per query.
    At training time, bipartite matching assigns predictions to the ground-truth set of subgraphs.%
  }
  \label{fig:architecture}
\end{figure*}

The model works as follows.
A graph neural network encoder produces per-node embeddings that retain the local structure of the input (Section~\ref{sec:encoder}). 
A fixed number~$Q$ of learnable \emph{query vectors} then interact with these node embeddings through a two-way transformer decoder, with each query predicting a specific subgraph instance (Section~\ref{sec:decoder}). 
A class head and a bilinear mask head decode each query into a (class, node-mask) pair. 
Queries that do not correspond to any instance predict a designated background class. 
The model is trained end-to-end with Hungarian matching, which assigns each prediction to a ground-truth instance, and a composite loss that penalizes both misclassification and mask inaccuracy (Section~\ref{sec:loss}). 
An overview of the architecture is visualized in Figure~\ref{fig:architecture}.

\subsection{Problem Formulation}

We formulate subgraph detection as a set prediction task over node-induced subgraph instances.
Let $G = (V, E, \BM{X})$ be a graph with $N=|V|$ nodes and node feature matrix $\BM{X} \in \mathbb{R}^{N \times d_x}$. %
A \emph{node-induced subgraph instance} is a pair $(c, \BM{m})$
consisting of a class label $c \in \{1, \ldots, C\}$ and a binary node mask $\BM{m} \in \{0, 1\}^{N}$.
A mask defines a node subset $V_{\BM{m}} = \{v \in V : \BM{m}_v = 1\}$, and the instance is the induced subgraph $G[V_{\BM{m}}] = (V_{\BM{m}},\, E_{\BM{m}})$ with $E_{\BM{m}} = \{(u, v) \in E : u, v \in V_{\BM{m}}\}$.
A graph $G$ may contain multiple instances of each pattern class, forming a set $\mathcal{S}^* = \{(c_k^*, \BM{m}_k^*)\}_{k=1}^{K}$, where $K$ varies across graphs and instances of the same or different classes may overlap in their node sets.
We note that this formulation also supports edge attributes (see Section~\ref{sec:experiments}) and easily extends to edge-induced subgraphs by replacing node masks with edge-membership masks.

The model is trained on a dataset of annotated graphs $\{(G_i, \mathcal{S}_i^*)\}_{i=1}^{n}$, where each $\mathcal{S}_i^*$ enumerates the instances present in $G_i$.
At inference time, given an unseen graph $G$, the goal is to predict a set $\hat{S}$ that matches the complete set $S^*$ of all instances.

\subsection{Graph Encoder}
\label{sec:encoder}

The encoder maps each node to an embedding that captures its features and structural context, producing a matrix $\BM{H} \in \mathbb{R}^{N \times D}$. 
Any graph neural network that produces per-node embeddings can serve as the backbone.
A vital aspect of the encoder is its expressive power, which directly influences the decoder's ability to recover subgraph instances. 
More precisely, if two nodes receive identical embeddings, the decoder cannot distinguish them and will produce identical mask scores for both across all queries, making it impossible to assign them to different instances.
Note that for nodes in the same orbit of the graph's automorphism group, no deterministic equivariant encoder can produce distinct embeddings. 
However, stochastic mechanisms such as random-walk sampling can break this symmetry, alleviating this issue.  
Complementary to discriminability, the encoder should produce representations that are structurally informative, encoding sufficient subgraph context for the decoder to determine pattern membership.
As one source of expressivity and structural signal, encoders optionally incorporate a Random Walk Structural Encoding (RWSE) \citep{dwivedi2022rwpe}, which enriches each node with random-walk return probabilities at increasing hop distances. 
We study three encoder families that differ in their expressivity and how they capture structural context.

\textbf{Message-passing neural networks (MPNNs).}
We consider GCN~\citep{kipf2017gcn}, GIN~\citep{xu2019gin}, and GatedGCN~\citep{bresson2017gatedgcn} equipped with the training recipes of~\citet{luo2025can}.
MPNNs iteratively aggregate neighborhood information. 
Their expressivity is bounded by the 1-WL test \citep{xu2019gin}, meaning they assign identical representations to nodes that the 1-WL algorithm fails to distinguish.

\textbf{Graph transformers.}
We use GraphGPS \citep{rampasek2022gps}, which combines local message passing with global self-attention, allowing each node to aggregate information beyond its immediate neighborhood.   
This allows each node to aggregate both local structural information and long-range context within each layer.
Despite this richer inductive bias, graph transformers remain permutation-symmetric in the standard formulation. 

\textbf{Random-walk neural networks.}
We employ NeuralWalker~\citep{chen2025neuralwalker} which combines random walks with message passing. It samples random walks from each node, encodes node and edge features along each walk with identity and adjacency encodings, and processes the resulting sequences with a sequence model. These structural encodings allow the model to identify induced subgraphs along walks, producing node representations that are informed by rich subgraph structure. Moreover, because each walk traces a unique trajectory through the graph, nodes at different positions receive distinct embeddings even when their local neighborhoods are isomorphic, breaking the symmetry that limits both MPNNs and standard graph transformers.

\subsection{Two-Way Transformer Decoder}
\label{sec:decoder}

The decoder transforms the per-node embeddings from the encoder into
$Q$ instance predictions, each consisting of a class label and a node-membership mask.
It maintains $Q$ learnable queries $\BM{Q} \in \mathbb{R}^{Q \times D}$ and node embeddings $\BM{H} \in \mathbb{R}^{N \times D}$, which are jointly updated by stacking two-way attention blocks, following the bidirectional decoder design of SAM \citep{kirillov2023sam, ravi2024sam2}.
A fixed positional encoding $\BM{E}_q$ is added to the queries at every layer.
We write one block as the following coupled update:
\begin{equation*}
\begin{aligned}
\BM{Q}^{\ell}_{\text{self}}
&= f_{\text{self}}\left(\BM{Q}^{\ell} + \BM{E}_q,
\BM{Q}^{\ell},
\BM{Q}^{\ell}\right), &
\BM{Q}^{\ell}_{\text{cross}}
&= f_{\text{cross}}\left(\BM{Q}^{\ell}_{\text{self}} + \BM{E}_q,\ \BM{H}^{\ell},\ \BM{H}^{\ell}\right), \\
\BM{Q}^{\ell+1}
&= f_{\text{ffn}} \left(\BM{Q}^{\ell}_{\text{cross}}\right), &
\BM{H}^{\ell+1}
&= f_{\text{rev}}\left(\BM{H}^{\ell},\ \BM{Q}^{\ell} + \BM{E}_q,\ \BM{Q}^{\ell}\right),
\end{aligned}
\end{equation*}
where $f_{\text{self}}, f_{\text{cross}}, f_{\text{rev}}$ denote multi-head attention modules (query–query, query–node, and node–query, respectively), and $f_{\text{ffn}}$ is a position-wise feedforward network. Standard LayerNorm and residual connections are applied to each sub-layer.

The update is bidirectional. 
Queries first interact via self-attention, then gather evidence from nodes, while nodes are updated in parallel by attending back to the queries.
The reverse attention induces query-conditioned node representations, allowing the mask head to assign different membership scores to the same node under different queries, which becomes necessary when instances overlap. %
After the final block, a standard cross-attention ($\BM{Q} \to \BM{H}$) produces $\BM{Q}_{\text{out}}$ and $\BM{H}_{\text{ref}}$, which are used by the class and mask heads, respectively.

\textbf{Class head.}
A linear projection maps each query to logits over $C+1$ classes, comprising $C$ foreground pattern classes and a background class $\varnothing$:
\begin{equation}
  \hat{\BM{l}}^{\text{cls}}_j
    = W_\text{cls}\, \BM{q}_j^\text{out}
    \in \mathbb{R}^{C+1}, \qquad j = 1, \ldots, Q,
\end{equation}
where $\BM{q}_j^\text{out}$ is the $j$-th query output from the final decoder block.

\textbf{Mask head.}
The mask head predicts node-membership logits through a bilinear interaction between a projected query and the query-conditioned node embeddings:
\begin{equation}
  \hat{\BM{l}}^{\text{mask}}_j
    = \BM{H}_\text{ref}\,
      (W_\text{mask}\, \BM{q}_j^\text{out})^\top
    \in \mathbb{R}^{N}.
\end{equation}
Using $\BM{H}_{\text{ref}}$ rather than the raw encoder output $\BM{H}$ is critical, because $\BM{H}_{\text{ref}}$ already encodes which queries are attending to which nodes.
Thus, the dot-product score measures how relevant each node is to query $j$'s current hypothesis rather than an unconditional node-query similarity. 
This gives the mask head access to information that flows in both directions of the decoder.

\subsection{Training Objective}
\label{sec:loss}

The decoder produces a fixed set of $Q$ predictions, while the number of ground-truth instances $K$ varies across graphs, with $K < Q$ in general.
Since there is no a priori assignment between predictions and ground-truth instances, training requires a matching step that pairs each ground-truth instance with exactly one prediction. 
We solve this via the Hungarian algorithm \citep{kuhn1955hungarian} and train with a composite loss on the matched pairs.

\textbf{Hungarian matching.}
For each graph, we determine an optimal matching $\sigma^*\colon [K] \to [Q]$ that assigns each ground-truth instance to a distinct query. 
This is achieved using the Hungarian algorithm applied to a cost matrix $\BM{C} \in \mathbb{R}^{Q \times K}$ that combines a classification term
and mask term:
\begin{equation}
  C_{j,k} =
      \underbrace{-\operatorname{softmax}
        (\hat{\BM{l}}^{\text{cls}}_j)_{c_k^*}}_{\text{class cost}}
    ~+~ \lambda_\text{mask}
      \underbrace{\mathcal{L}_\text{BCE}
        (\hat{\BM{l}}^{\text{mask}}_j,\,\BM{m}_k^*)}_{\text{mask cost}}
\end{equation}
Since the matching is non-differentiable, gradients flow exclusively through the training loss evaluated on the matched pairs.
Unmatched queries $j \notin \sigma^*([K])$ are assigned the background class.

\textbf{Loss function.}
Given the matching $\sigma^*$, the total training loss is given by:
\begin{equation}
  \mathcal{L}
    = \mathcal{L}_\text{CE}
    + \lambda_\text{mask}\,\mathcal{L}_\text{mask}
\end{equation}
where $\mathcal{L}_\text{CE}$ is a cross-entropy loss over all $Q$ query slots. 
Unmatched slots target class $\varnothing$ and are down-weighted by $w_\varnothing < 1$ to counterbalance the imbalance between the few matched queries and the many background ones. 
$\mathcal{L}_\text{mask}$ is applied only to matched pairs:
\begin{equation}
  \mathcal{L}_\text{mask}
    = \mathcal{L}_\text{BCE}
    + \lambda_\text{cut}\,\mathcal{L}_\text{cut}.
\end{equation}
The binary cross-entropy term $\mathcal{L}_\text{BCE}$ provides per-node supervision.

\textbf{Graph cut penalty.}
The term $\mathcal{L}_\text{cut}$ encodes a structural prior that subgraph instances are connected subgraphs, penalizing edges that cross the predicted membership boundary.
For a matched pair, the penalty is computed from the predicted membership probabilities $p_u = \sigma(l_u)$ as follows:
\begin{equation}
  \mathcal{L}_\text{cut}
    = \frac{1}{|\mathcal{E}|}
      \sum_{(u,v)\in\mathcal{E}}
      \bigl[p_u(1-p_v)+p_v(1-p_u)\bigr].
\end{equation}
Each summand is zero when both endpoints are confidently on the same side of the boundary ($p\approx 0$ or $p\approx 1$) and reaches $\nicefrac{1}{2}$ when they lie on opposite sides.
The gradient drives uncertain boundary nodes toward the majority vote of their neighbors, pushing a node inward if most neighbors are predicted inside the subgraph, and outward otherwise (see Appendix~\ref{app:cut} for a full derivation).
This implements a differentiable relaxation of the subgraph connectivity constraint without requiring any auxiliary connected-component check during training.

\section{Molecular Functional Group Detection}
\label{sec:experiments}

We apply \GraphDETR{} to molecular functional group (FG) detection, an instance-level subgraph detection problem where both the class and the exact atom-membership mask of every FG instance need to be jointly recovered.
We evaluate \GraphDETR{} on ChEMBL\footnote{\url{https://chembl.gitbook.io/}}, drawn from ChEMBL-36, containing $2{,}827{,}875$ molecules across $95$ functional group classes.
For ablation studies we use ChEMBL12k, a subset of $12{,}000$ molecules. 
Full dataset details can be found in Appendix~\ref{app:datasets}.

\textbf{Evaluation metrics.}
We adopt an object detection evaluation protocol adapted to graphs. Each predicted node-set mask is matched to a ground-truth instance by Hungarian assignment, using the Intersection over Union (IoU) of their node sets, i.e., $|V_\text{pred} \cap V_\text{gt}|\,/\,|V_\text{pred} \cup V_\text{gt}|$, as the similarity measure.
We report five metrics:
$\textbf{AP}_{100}$ (average precision at IoU $= 1.0$, macro-averaged over classes),
$\textbf{mAP}$ (COCO-style, thresholds $0.50$--$0.95$ in steps of $0.05$),
$\textbf{Rec}_{100}$ (recall at IoU $= 1.0$),
$\textbf{mIoU}$ (mean mask IoU over matched true-positive pairs), and
$\textbf{ExactMatch}$ (fraction of molecules where every GT instance is matched with IoU $\geq 0.5$ and no false positive is active).

\subsection{Graph Encoder Comparison}
\label{sec:encoder_comparison}

\begin{table}[t]
  \centering
  \caption{%
    \textbf{Functional group detection on ChEMBL12k.}
    All \GraphDETR{} models use $Q{=}40$ queries, $5$ encoder layers, $D{=}256$, RWSE ($K{=}20$),
    and the full loss.
    NeuralWalker uses the Mamba sequence backend.
    Results are mean\,$\pm$\,std over 5 seeds.
  }
  \label{tab:main}
  \small
  \resizebox{\textwidth}{!}{
  \begin{tabular}{llcccccc}
    \toprule
    \textbf{Encoder} & \textbf{\# Params}
      & \textbf{AP$_{100}$} & \textbf{mAP$_{50:95}$}
      & \textbf{Rec$_{100}$} & \textbf{mIoU}
      & \textbf{ExactMatch} \\
    \midrule
    GCN       & 3.9M & 0.8669$_{\pm 0.0122}$ & 0.8766$_{\pm 0.0106}$ & 0.8631$_{\pm 0.0082}$ & 0.9768$_{\pm 0.0025}$ & 0.7424$_{\pm 0.0273}$ \\
    GIN       & 4.9M & 0.8815$_{\pm 0.0104}$ & 0.8888$_{\pm 0.0100}$ & 0.8724$_{\pm 0.0043}$ & 0.9792$_{\pm 0.0037}$ & 0.7642$_{\pm 0.0153}$ \\
    GatedGCN  & 5.3M & 0.8687$_{\pm 0.0078}$ & 0.8751$_{\pm 0.0072}$ & 0.8735$_{\pm 0.0032}$ & 0.9756$_{\pm 0.0028}$ & 0.5948$_{\pm 0.0259}$ \\
    GraphGPS  & 6.6M & 0.9191$_{\pm 0.0112}$ & 0.9230$_{\pm 0.0099}$ & 0.8923$_{\pm 0.0017}$ & 0.9812$_{\pm 0.0012}$ & 0.7460$_{\pm 0.0286}$ \\
    NeuralWalker ($\lambda_\text{mask}{=}2$) & 7.4M & 0.9136$_{\pm 0.0165}$ & 0.9207$_{\pm 0.0164}$ & 0.9567$_{\pm 0.0102}$ & 0.9912$_{\pm 0.0036}$ & \textbf{0.8912}$_{\pm 0.0264}$ \\
    NeuralWalker ($\lambda_\text{mask}{=}4$) & 7.4M & \textbf{0.9370}$_{\pm 0.0147}$ & \textbf{0.9438}$_{\pm 0.0135}$ & \textbf{0.9718}$_{\pm 0.0037}$ & \textbf{0.9944}$_{\pm 0.0009}$ & 0.8342$_{\pm 0.0320}$ \\
    \bottomrule
  \end{tabular}
  }
\end{table}

We start by comparing the encoder backbones introduced in Section~\ref{sec:encoder}, keeping all other components of \GraphDETR{} fixed. 
As shown in Table~\ref{tab:main}, \GraphDETR{} achieves strong detection performance across all encoders, demonstrating that the set prediction formulation is effective for this combinatorially rich real-world task. 
Notably, local MPNN encoders (GCN, GIN, GatedGCN) are consistently outperformed by GraphGPS and NeuralWalker across all precision-oriented metrics, with NeuralWalker achieving the best overall performance.
This likely stems from the stronger ability of GraphGPS and NeuralWalker to capture global structural context, in contrast to the purely local aggregation of MPNNs.

Among local encoders, GatedGCN achieves competitive AP$_{100}$ but its ExactMatch (i.e., the ability to find all instances exactly) is substantially lower than both GCN and GIN, suggesting that it struggles to cleanly separate co-occurring instances at the molecule level.
Furthermore, it is notable that mIoU is uniformly high across all encoders, which shows that matched pairs are well localized regardless of backbone.
Differences between encoders are therefore most pronounced in recall and ExactMatch rather than in the quality of masks for already-identified instances.

\textbf{Data scaling (Figure~\ref{fig:scaling}).}
We further evaluate how detection performance scales with training data size on the full ChEMBL dataset.
Both GIN and NeuralWalker improve log-linearly as the number of training molecules grows, with NeuralWalker consistently ahead.
The parallel slopes suggest that NeuralWalker's advantage over GIN does not erode with scale, pointing to a structural rather than a data-driven benefit.

\subsection{Ablation and Hyperparameter Studies}
\label{sec:ablation}

In this section, we ablate key components of \GraphDETR{} and study the sensitivity to central hyperparameters on the ChEMBL12k dataset.
Specifically, we examine the contribution of each loss term, the impact of the decoder capacity, and the sensitivity of the NeuralWalker encoder to its hyperparameter choices.
Qualitative predictions on held-out test molecules are provided in Appendix~\ref{app:results}.

\begin{figure}[t]
  \centering
  \begin{minipage}[t]{0.47\textwidth}
    \centering
    \captionof{table}{%
      \textbf{Loss ablation (OFAT)} on the ChEMBL12k dataset.
      Each row adds one loss component relative to the previous.%
    }
    \label{tab:loss_abl}
    \resizebox{\linewidth}{!}{%
    \begin{tabular}{lccc}
      \toprule
      Loss & AP$_{100}$ & ExactMatch & Rec$_{100}$ \\
      \midrule
      CE only        &  0.4 &  0.8 &  0.5 \\
      $+$ BCE        & 85.4 & 79.1 & 95.0 \\
      $+$ BG weight  & 92.4 & 84.8 & 97.1 \\
      Full ($+$ cut) & 94.3 & 81.8 & 97.2 \\
      \bottomrule
    \end{tabular}}
  \end{minipage}
  \hfill
  \begin{minipage}[t]{0.47\textwidth}
    \centering
    \captionof{figure}{%
      \textbf{Data scaling} on ChEMBL (AP$_{100}$, test set).
  Encoders improve log-linearly with scale,
  with NeuralWalker consistently ahead.%
    }
    \label{fig:scaling}
    \vspace{-5pt}
    \begin{tikzpicture}
\begin{axis}[
  xmode         = log,
  width         = 0.75\textwidth,
  height        = 3.2cm,
  xlabel        = {\scriptsize Training molecules},
  ylabel        = {\scriptsize AP$_{100}$ (\%)},
  y label style = {at={(-0.15,0.5)}},
  xlabel style = {yshift=4pt},
  ylabel style  = {font=\small},
  xmin          = 10000, xmax = 10000000,
  ymin          = 70,  ymax = 100,
  xtick         = {10000,100000,1000000,10000000},
  xticklabels   = {$10^4$,$10^5$,$10^6$,$10^7$},
  ymajorgrids   = true,
  xmajorgrids   = true,
  grid style    = {gray!25},
  tick label style    = {font=\scriptsize},
  x tick label style  = {font=\scriptsize},
  legend style  = {at={(0.99,0.03)}, anchor=south east,
                         font=\scriptsize, cells={anchor=west},
                         row sep=0pt, inner sep=0pt},
        legend pos=outer north east
]
\addplot[color=blockblue, mark=triangle*, mark size=1.8pt, thick]
  coordinates {
    (22623,   80.69)
    (226230,  84.85)
    (2262300, 87.31)
  };
\addlegendentry{GIN}
\node[font=\tiny, color=blockblue, anchor=north west] at (axis cs:22623, 80.69) {80.7};
\node[font=\tiny, color=blockblue, anchor=north]      at (axis cs:226230,  84.85) {84.9};
\node[font=\tiny, color=blockblue, anchor=north east] at (axis cs:2262300, 87.31) {87.3};
\addplot[color=blockorange, mark=*, mark size=1.8pt, thick]
  coordinates {
    (22623,   86.61)
    (226230,  87.53)
    (2262300, 91.18)
  };
\addlegendentry{NeuralWalker}
\node[font=\tiny, color=blockorange, anchor=south west] at (axis cs:22623,   86.61) {86.6};
\node[font=\tiny, color=blockorange, anchor=south]      at (axis cs:226230,  87.53) {87.5};
\node[font=\tiny, color=blockorange, anchor=south west] at (axis cs:2262300, 91.18) {91.2};
\end{axis}
\end{tikzpicture}
  \end{minipage}
  \vspace{-20pt}
\end{figure}

\textbf{Loss components (Table~\ref{tab:loss_abl}).}
We ablate the training objective introduced in Section~\ref{sec:loss} by progressively adding each loss term.
Without mask supervision, the model fails to localize instances entirely, confirming that cross-entropy supervision (CE) on class predictions alone is insufficient for instance-level detection. 
Adding the binary cross-entropy mask loss (BCE) recovers most of the performance, and down-weighting the background class (BCE+BG weight) further reduces false positive predictions substantially. 
Finally, the full objective, which additionally includes the graph cut penalty (Full+cut), further improves AP$_{100}$ and Rec$_{100}$, though at a slight cost to ExactMatch.

\begin{table}[t]
  \centering
  \caption{%
    \textbf{Decoder ablation (OFAT)} on the ChEMBL12k validation set.
    Baseline: NeuralWalker encoder with \# layers${=}3$, \# queries${=}40$, $h{=}8$ heads, attn downsample rate $r{=}2$. Each row varies exactly one axis.%
  }
  \label{tab:decoder}
  \small
  \resizebox{.74\textwidth}{!}{
  \begin{tabular}{lcccccc}
    \toprule
    \textbf{Varied axis} & \textbf{Value}
      & \textbf{AP$_{100}$} & \textbf{mAP} & \textbf{Rec$_{100}$} & \textbf{mIoU} & \textbf{ExactMatch} \\
    \midrule
    baseline & -- & \textbf{0.9429} & \textbf{0.9497} & 0.9720 & 0.9939 & 0.8180 \\
    \midrule
    \# layers & 1  & 0.8983 & 0.9050 & 0.9493 & 0.9885 & 0.8100 \\
    \# layers & 2  & 0.9207 & 0.9273 & 0.9690 & 0.9936 & 0.8460 \\
    \midrule
    \# queries & 10 & 0.8674 & 0.8822 & 0.9107 & 0.9858 & 0.7080 \\
    \# queries & 20 & 0.9296 & 0.9371 & 0.9563 & 0.9924 & 0.8180 \\
    \# queries & 30 & 0.9166 & 0.9247 & 0.9635 & 0.9925 & 0.8420 \\
    \midrule
    \# attn heads        & 4  & 0.8953 & 0.9064 & 0.9502 & 0.9893 & 0.7020 \\
    \midrule
    attn downsample rate & 1 & 0.9062 & 0.9169 & 0.9639 & 0.9923 & 0.8500 \\
    attn downsample rate & 4 & 0.9312 & 0.9411 & \textbf{0.9739} & \textbf{0.9949} & \textbf{0.9000} \\
    \bottomrule
  \end{tabular}
  }
  \vspace{-10pt}
\end{table}

\textbf{Decoder capacity (Table~\ref{tab:decoder}).}
We study the effect of the main architectural hyperparameters of the two-way transformer decoder (Section~\ref{sec:decoder}), \ie the number of attention blocks, query slots, and attention heads.
Furthermore, we analyze the effect of the downsample rate $r$, which compresses the attention key/value dimensions to $D/r$.
Reducing the number of attention blocks degrades AP$_{100}$ gradually, suggesting that decoder depth contributes meaningfully to detection performance. 
Concerning the number of query slots $Q$, it can be observed that $Q{=}10$ leads to a sharp performance drop, possibly because many molecules contain more than 10 functional group instances, leaving the decoder unable to cover them all simultaneously. 
The rather gradual degradation observed at higher values of $Q$ may reflect the decreasing fraction of molecules affected as $Q$ grows.
Turning to the number of attention heads, we find that halving from 8 to 4 causes a larger performance drop than any reduction in depth, showing that multi-head diversity is particularly important for decoder performance.
Finally, increasing the attention downsample rate $r$ improves ExactMatch notably at a mild cost to AP$_{100}$, possibly suggesting that compressing the key/value space acts as a form of regularization that promotes more decisive mask predictions.

\begin{table}[t]
  \centering
  \caption{%
    \textbf{NeuralWalker encoder ablation (OFAT)} on the ChEMBL12k validation
    set.%
  }
  \label{tab:encoder}
  \small
  \resizebox{.95\textwidth}{!}{
  \begin{tabular}{lccccccccc}
    \toprule
    \textbf{Varied axis} & \textbf{Value}
      & \textbf{Layers $L$} & \textbf{Walk $\ell$}
       & \textbf{RWSE}
      & \textbf{AP$_{100}$} & \textbf{mAP} & \textbf{Rec$_{100}$} & \textbf{mIoU} & \textbf{ExactMatch} \\
    \midrule
    baseline   & --   & 5 & 20  & \cmark ($K{=}20$) & \textbf{0.9429} & \textbf{0.9497} & \textbf{0.9720} & \textbf{0.9939} & 0.8180 \\
    \midrule
    \# layers & 2   & 2 & 20  & \cmark & 0.9122 & 0.9230 & 0.9617 & 0.9910 & 0.8380 \\
    \# layers & 3   & 3 & 20  & \cmark & 0.9130 & 0.9219 & 0.9688 & 0.9930 & \textbf{0.8530} \\
    \# layers & 4   & 4 & 20  & \cmark & 0.9352 & 0.9428 & 0.9715 & 0.9931 & 0.8480 \\
    \midrule
    walk length & 10 & 5 & 10  & \cmark & 0.9292 & 0.9344 & 0.9581 & 0.9918 & 0.8500 \\
    walk length & 40 & 5 & 40  & \cmark & 0.8931 & 0.9021 & 0.9625 & 0.9931 & 0.8540 \\
    \midrule
    RWSE & \xmark  & 5 & 20  & \xmark & 0.9225 & 0.9292 & 0.9635 & 0.9921 & 0.8480 \\
    \midrule
    edge attr & off & 5 & 20 & \cmark & 0.9170 & 0.9246 & 0.9678 & 0.9921 & 0.8240 \\
    \bottomrule
  \end{tabular}
  }
  \vspace{-15pt}
\end{table}

\textbf{NeuralWalker encoder sensitivity (Table~\ref{tab:encoder}).}
Given that NeuralWalker achieves the best overall performance in Section~\ref{sec:encoder_comparison}, we study the sensitivity of its key parameter choices, i.e., the number of NeuralWalker blocks, walk length, and the use of RWSE and edge features. 
The number of NeuralWalker blocks $L$ has only a modest effect on performance across the tested values, with AP$_{100}$ improving gradually with depth while ExactMatch peaks at an intermediate value and then slightly declines.
Turning to walk length, an optimum is observed at $\ell{=}20$, with shorter walks providing insufficient structural context and, interestingly, longer walks actively hurting performance.
Removing RWSE, which enriches node features with random-walk return probabilities, leads to a meaningful drop in AP$_{100}$, suggesting that it provides further structural information beyond what the walk sequences themselves capture.
Finally, ablating edge features also leads to a meaningful drop in AP$_{100}$, confirming that edge attributes provide further useful structural signal for instance detection.

\section{General Subgraph Detection Capabilities}
\label{sec:synthetic}

To assess the generality of \GraphDETR{} beyond molecular graphs, we evaluate it on a diverse set of subgraph detection benchmarks covering exact and approximate pattern matching across a wide range of graph sizes and pattern types. 
In addition, we provide inference runtimes and study the ability of trained models to generalize to larger graphs than those seen during training.

We evaluate on five dataset families: Cactus graphs in which cycles of varying lengths are the query patterns, bipartite graphs with injected $k$-cliques as query patterns, ZINC12k~\citep{dwivedi2023benchmarking} where frequent molecular subgraphs serve as query patterns, Mol-Reddit where molecular structures injected into social network host graphs are the targets of detection, and fuzzy variants of the Cactus and Clique datasets where instances only approximately match the query pattern.
For the Cactus and Clique datasets, the graph generation procedure is designed to prevent accidental pattern occurrences, ensuring complete ground-truth annotations. For ZINC12k, completeness is guaranteed by the exhaustive frequent subgraph enumeration. 
Full details are provided in Appendix~\ref{app:datasets}.
We report AP$_{100}$ as the primary metric, following the same evaluation protocol as in Section~\ref{sec:experiments}. 
Model configuration follows Section~\ref{sec:experiments}, with hyperparameters listed in Appendix~\ref{app:sec:hparams}.

\textbf{Exact detection (Table~\ref{tab:synthetic}).}
We start by examining exact subgraph detection across the synthetic benchmarks.
As in Section~\ref{sec:encoder_comparison}, NeuralWalker and GraphGPS consistently outperform local MPNN encoders across all datasets, confirming that global structural context is beneficial beyond the molecular setting.
Performance on Cactus graphs degrades substantially with graph size, which is expected given that both the host graphs and the cycle patterns grow larger, making disambiguation increasingly difficult.
In contrast, performance on Clique graphs remains strong even as the clique size grows, suggesting that the dense connectivity of cliques provides a distinctive structural signature that is easier to detect regardless of size.
ZINC12k and Mol-Reddit present a different challenge as they involve attributed graphs with molecular (sub-)structures as patterns. Notably, NeuralWalker performs particularly strongly on Mol-Reddit, which is arguably the most challenging dataset due to its incomplete annotations and large, varied host graphs. 

\textbf{Approximate detection (Table~\ref{tab:synthetic}).}
We further evaluate \GraphDETR{} on approximate pattern matching, where instances are defined by their proximity to a canonical query pattern under graph edit distance (GED). 
The task requires separating instances whose GED to the canonical pattern falls below a threshold from those that do not, which is challenging because GED computation is itself NP-hard.
Despite this difficulty, \GraphDETR{} achieves surprisingly strong performance on both fuzzy datasets, demonstrating that it can learn to implicitly discriminate between structurally similar subgraphs purely from data, without any explicit GED computation.
This is particularly striking as the model must correctly reject subgraphs that differ from the query pattern by only a small number of edits, a fine-grained structural distinction that emerges purely from training. 

\begin{table}[tbp]
  \centering
  \caption{%
    \textbf{Synthetic subgraph detection} in AP$_{100}$\%. $n$/$m$ denotes the average number of nodes/edges.
  }\label{tab:synthetic}
  \small
  \resizebox{\textwidth}{!}{
  \begin{tabular}{llllccccc}
    \toprule
    \multirow{2}{*}{\textbf{Dataset}}
    & \multirow{2}{*}{\textbf{Patterns (\#)}}
    & \multirow{2}{*}{\textbf{Attr.}}
    & \multirow{2}{*}{\textbf{Stats}}
    & \multicolumn{4}{c}{\textbf{\GraphDETR{} encoder}} \\ \cmidrule(lr){5-8}
    & & & & GCN
    & GIN 
    & GraphGPS 
    & NeuralWalker \\
    \midrule
    Cactus40
    & (3-8)-cycles (6)
    & \xmark
    & $n=40,m=45$
    & $92.33_{\pm 0.56}$ 
    & $92.98_{\pm 0.90}$ 
    & $96.51_{\pm 0.43}$ 
    & $\textbf{97.44}_{\pm 0.44}$ \\

    Cactus200
    & (11-16)-cycles (6)
    & \xmark
    & $n=200,m=204$
    & $42.80_{\pm 3.16}$ 
    & $41.51_{\pm 4.45}$ 
    & $83.67_{\pm 0.78}$ 
    & $\textbf{91.90}_{\pm 2.41}$ \\

    Cactus1000 
    & (27-32)-cycles (6)
    & \xmark
    & $n=1000,m=1004$
    & $10.46_{\pm 3.04}$ 
    & $4.82_{\pm 7.30}$ 
    & $40.22_{\pm 0.68}$ 
    & $\textbf{42.20}_{\pm 2.06}$ \\

    Cliques100 
    & 5-clique (1)
    & \xmark
    & $n=140,m=350$
    & $90.94_{\pm 2.48}$ 
    & $93.31_{\pm 0.74}$ 
    & $\textbf{97.05}_{\pm 1.60}$ 
    & $96.15_{\pm 0.57}$ \\

    Cliques200
    & 10-clique (1)
    & \xmark
    & $n=290,m=1450$
    & $90.33_{\pm 1.46}$ 
    & $92.84_{\pm 2.20}$ 
    & $\textbf{96.45}_{\pm 1.17}$ 
    & $93.48_{\pm 2.35}$ \\

    Cliques400
    & 20-clique (1)
    & \xmark
    & $n=590,m=5900$
    & $92.51_{\pm 2.85}$ 
    & $94.12_{\pm 1.51}$ 
    & $\textbf{95.05}_{\pm 3.23}$
    & $92.58_{\pm 3.48}$ \\

    ZINC12k 
    & frequent subgraphs (9)
    & \cmark
    & $n=23,m=25$
    & $89.68_{\pm 0.88}$ 
    & $92.86_{\pm 0.25}$ 
    & $95.28_{\pm 0.13}$ 
    & $\textbf{97.66}_{\pm 0.27}$ \\

    Mol-Reddit 
    & injected molecules (10)
    & \xmark
    & $n=441, m=517$
    & $65.68_{\pm 1.11}$ 
    & $62.12_{\pm 1.31}$ 
    & $80.78_{\pm 1.32}$ 
    & $\textbf{89.76}_{\pm 0.97}$ \\ \midrule
    Cactus-fuzzy
    & fuzzy (7-10)-cycles (4)
    & \cmark
    & $n=100, m=105$
    & $67.78_{\pm 1.17}$
    & $76.19_{\pm 2.03}$
    & $84.86_{\pm 0.51}$
    & $\textbf{90.91}_{\pm 2.88}$\\

    Cliques-fuzzy
    & fuzzy 8-clique (1)
    & \xmark
    & $n=170, m=470$
    & $95.05_{\pm 1.02}$
    & $95.26_{\pm 0.89}$
    & $\textbf{96.66}_{\pm 0.86}$
    & $93.94_{\pm 2.10}$ \\
    \bottomrule
  \end{tabular}
  }
\end{table}

\textbf{Runtime (Table~\ref{tab:runtime}).}
We compare the inference runtime of \GraphDETR{} against VF2~\citep{cordella2004sub}, which is the de facto standard algorithm for subgraph isomorphism, a highly optimized combinatorial baseline.
Nevertheless, VF2 fails to terminate within the provided time budget on a large amount of test graphs across several datasets, while \GraphDETR{} processes all graphs in a single forward pass at runtimes several orders of magnitude lower.
This gap widens with graph size, consistent with the exponential worst-case complexity of combinatorial subgraph isomorphism search, while \GraphDETR{} maintains low runtimes even on the largest graphs.

\textbf{Generalization to larger graphs (Figure~\ref{fig:transfer}).}
We evaluate the ability of a single model trained on small graphs (Cactus40 and Cliques100) to generalize to increasingly larger graphs at test time.
For cycles, all encoders significantly degrade as graph size grows, with NeuralWalker and GraphGPS degrading slightly slower than MPNNs.
For cliques, the picture is more nuanced and somewhat surprising.
GraphGPS proves to be the most robust encoder, maintaining strong performance well beyond the training distribution, while NeuralWalker degrades more significantly at larger graph sizes despite being the best encoder in-distribution.

\begin{figure}[tp]
  \centering
  \vspace{-5pt}
  \begin{minipage}[t]{0.48\linewidth}
    \captionof{table}{%
      \textbf{Inference runtime (s\,/\,graph).}
      VF2 uses a 60\,s per-graph budget; $^\dag$ marks datasets where
      VF2 timed out on $>$80\% of test graphs.
      \GraphDETR{} runtimes are means across seeds.%
    }
    \label{tab:runtime}
    \vspace{-3pt}
    \centering
    \small
    \resizebox{\linewidth}{!}{%
    \begin{tabular}{lrrrrr}
      \toprule
      \multirow{2}{*}{\textbf{Dataset}} & & \multicolumn{4}{c}{\textbf{GraphDETR}} \\ \cmidrule(lr){3-6}
        & VF2~\citep{cordella2004sub} & GCN & GIN & GPS & NW \\
      \midrule
      Cactus40      & 0.068     & 5.4e-5 & 5.2e-5 & 9.8e-5 & 0.0014 \\
      Cactus200     & 2.799     & 9.3e-5 & 9.2e-5 & 0.0002 & 0.0174 \\
      Cactus1000    & 60$^\dag$ & 0.0002 & 0.0002 & 0.0009 & 0.0412 \\
      Cliques100    & 0.120     & 5.8e-5 & 6.2e-5 & 0.0001 & 0.0041 \\
      Cliques200    & 60$^\dag$ & 8.5e-5 & 9.0e-5 & 0.0003 & 0.0038 \\
      Cliques400    & 60$^\dag$ & 0.0002 & 0.0002 & 0.0007 & 0.0079 \\
      ZINC12k       & 0.010     & 5.5e-5 & 5.3e-5 & 9.9e-5 & 0.0021 \\
      Mol-Reddit    & 51$^\dag$ & 0.0002 & 0.0002 & 0.0021 & 0.0125 \\
      Cactus-fuzzy  & --        & 5.6e-5 & 5.6e-5 & 0.0001 & 0.0045 \\
      Cliques-fuzzy & --        & 6.1e-5 & 6.6e-5 & 0.0002 & 0.0043 \\
      \bottomrule
    \end{tabular}}
  \end{minipage}%
  \hfill%
  \begin{minipage}[t]{0.49\linewidth}
    \captionof{figure}{%
  \textbf{Generalization to larger graphs.}
  AP$_{100}$ of a single model trained on small graphs,
  evaluated at increasing graph sizes.
  Left: trained on Cactus40. Right: trained on Cliques100.%
    }
    \label{fig:transfer}
    \vspace{-5pt}
    \centering
    \definecolor{myTeal}{HTML}{2E9B8F}
\definecolor{myPurple}{HTML}{7B5EA7}
\definecolor{myBlue}{HTML}{3A7EC6}
\definecolor{myGold}{HTML}{C89B2A}
\definecolor{myRose}{HTML}{C05060}

\begin{tikzpicture}
  \begin{groupplot}[
    group style={
      group size        = 2 by 1,
      horizontal sep    = 0.2cm,
    },
    width               = 0.655\linewidth,
    height              = 4.0cm,
    ymin=0, ymax=100,
    ymajorgrids         = true,
    grid style          = {gray!20},
    tick label style    = {font=\tiny},
    ylabel style        = {font=\scriptsize, at={(-0.11,0.5)}},
    xlabel style        = {font=\scriptsize},
    title style         = {font=\scriptsize, yshift=-4pt},
    legend style        = {
      at={(0.95,-0.32)},
      anchor=north,
      legend columns=4,
      font=\scriptsize,
      cells={anchor=west},
      column sep=0pt,
      inner sep=0pt,
    },
  ]
  \nextgroupplot[
    xtick           = {40,50,60,70,80,90,100},
    xmin=35, xmax=105,
    xlabel style = {font=\scriptsize, yshift=4pt},
    xlabel          = {Graph size},
    ylabel          = {AP$_{100}$ (\%)},
  ]
  \addplot[color=myRose, mark=diamond*, mark size=1.5pt, line width=0.8pt]
    coordinates {(40,92.3)(50,75.9)(60,55.4)(70,40.3)(80,30.1)(90,23.1)(100,17.9)};
  \addlegendentry{GCN}
  \addplot[color=myGold, mark=triangle*, mark size=1.5pt, line width=0.8pt]
    coordinates {(40,93.0)(50,75.5)(60,53.6)(70,39.8)(80,30.1)(90,23.1)(100,18.3)};
  \addlegendentry{GIN}
  \addplot[color=myBlue, mark=square*, mark size=1.5pt, line width=0.8pt]
    coordinates {(40,96.5)(50,85.1)(60,62.4)(70,41.7)(80,28.4)(90,19.2)(100,13.8)};
  \addlegendentry{GraphGPS}
  \addplot[color=myTeal, mark=*, mark size=1.5pt, line width=0.8pt]
    coordinates {(40,97.4)(50,86.9)(60,63.9)(70,44.1)(80,30.5)(90,20.9)(100,14.9)};
  \addlegendentry{NeuralWalker}
  \nextgroupplot[
    xtick           = {140,160,180,200,220,240},
    xmin=130, xmax=250,
    xlabel style = {font=\scriptsize, yshift=4pt},
    xlabel          = {Graph size},
    yticklabels     = {},
  ]
  \addplot[color=myTeal, mark=*, mark size=1.5pt, line width=0.8pt]
    coordinates {(140,96.2)(160,83.3)(180,27.5)(200,8.3)(220,1.4)(240,0.2)};
  \addplot[color=myBlue, mark=square*, mark size=1.5pt, line width=0.8pt]
    coordinates {(140,97.1)(160,95.5)(180,87.7)(200,69.0)(220,50.0)(240,40.7)};
  \addplot[color=myGold, mark=triangle*, mark size=1.5pt, line width=0.8pt]
    coordinates {(140,93.3)(160,86.4)(180,57.1)(200,30.3)(220,14.9)(240,7.4)};
  \addplot[color=myRose, mark=diamond*, mark size=1.5pt, line width=0.8pt]
    coordinates {(140,90.9)(160,81.5)(180,54.9)(200,24.8)(220,7.6)(240,2.5)};
  \end{groupplot}
\end{tikzpicture}
  \end{minipage}
  \vspace{-20pt}
\end{figure}

\section{Conclusion}
\label{sec:conclusion}

This work introduced \GraphDETR{}, an end-to-end framework for subgraph detection that casts the problem as set prediction. By combining graph encoders with a two-way transformer decoder and bipartite matching, the model predicts the complete set of subgraph instances in a single forward pass, eliminating the need for combinatorial search or post-processing. This formulation naturally supports both exact and approximate matching, extending beyond the rigid constraints of classical subgraph isomorphism.

Empirical results across molecular and synthetic benchmarks demonstrate that \GraphDETR{} achieves strong detection performance, scales to large graphs, and operates with orders-of-magnitude faster inference than traditional algorithms. The analysis further highlights the importance of expressive graph encoders and the role of global structural context in instance-level graph reasoning.

More broadly, the proposed set prediction perspective provides a general and flexible framework for structured prediction on graphs and a useful framework for evaluating expressive graph encoders. Future work may explore richer query conditioning, improved scalability to even larger graphs, and extensions to other structured outputs such as edge-induced subgraphs or hierarchical patterns.

\begin{ack}
The authors would like to thank Cheng-Wei Liao for his insightful discussions, which have inspired this work.
\end{ack}

\bibliographystyle{plainnat}
\bibliography{refs}

\newpage
\appendix
\vspace*{0.3cm}
\begin{center}
    {\huge Appendix}
\end{center}
\vspace*{0.5cm}

The appendix provides supplementary technical and experimental materials. It is organized as follows: Section~\ref{app:sec:limitations} discusses the limitations of the proposed approach. Section~\ref{app:sec:broader_impacts} provides a discussion on the broader impacts. Section~\ref{app:sec:llms} lists the use of large language models. Section~\ref{app:cut} provides the derivation and analysis of the graph cut penalty used in the training objective. Section~\ref{app:datasets} provides details of the datasets used in this work. Section~\ref{app:impl} provides full implementations for reproducibility purposes. Section~\ref{app:results} provides additional experimental results.

\section{Limitations}\label{app:sec:limitations}

While \GraphDETR{} demonstrates strong empirical performance and scalability, several limitations remain.

First, the approach relies on a fixed number of query slots, which imposes an upper bound on the number of detectable instances per graph. Although this can be mitigated by choosing a sufficiently large number of queries, doing so increases computational cost and may introduce optimization challenges due to a larger proportion of unmatched queries.

Second, the model’s performance depends critically on the expressivity of the graph encoder. As discussed in Section~\ref{sec:encoder}, encoders with limited discriminative power (\eg standard MPNNs) may fail to distinguish structurally symmetric nodes, which can hinder instance separation. While more expressive architectures, such as Graph Transformers or random-walk-based models, alleviate this issue, they may incur higher computational overhead, and no single model dominates on all datasets.

Third, although \GraphDETR{} scales well compared to combinatorial methods, its memory and runtime still grow with graph size due to attention operations in the decoder. This may limit applicability to extremely large graphs without further architectural modifications, such as sparse or hierarchical attention.

Fourth, the current formulation is limited to a closed vocabulary of pattern classes observed during training. The model cannot directly generalize to detect entirely novel or unseen subgraph patterns at inference time, in contrast to open-vocabulary detection settings explored in other domains. Enabling open-vocabulary or query-conditioned subgraph detection, where patterns are specified at test time, remains an important direction for future work.

Finally, the current formulation assumes access to fully annotated training data with instance-level supervision (i.e., node masks). In many real-world settings, such annotations are costly or unavailable. Extending the framework to weakly supervised or unsupervised settings remains an open challenge.

\section{Broader Impacts}\label{app:sec:broader_impacts}
\GraphDETR{} provides a general framework for detecting structured patterns in graphs, with potential applications across chemistry, biology, and network science.

In molecular science, the ability to accurately detect functional groups and substructures may accelerate drug discovery, retrosynthesis planning, and materials design. By enabling scalable and flexible pattern detection, \GraphDETR{} could support the identification of novel chemical motifs and improve predictive modeling pipelines.

In network analysis, the method may facilitate the discovery of recurring structural motifs in social, biological, or technological networks, contributing to a deeper understanding of their organization and function.

However, these capabilities also raise potential concerns. In chemistry and biology, improved substructure detection could be misused to design harmful compounds or bypass safety constraints in molecular screening pipelines. In social network analysis, identifying structural patterns at scale may raise privacy concerns if applied to sensitive relational data.

Mitigating these risks requires responsible deployment, including adherence to domain-specific regulations, careful dataset curation, and consideration of ethical implications. We emphasize that \GraphDETR{} is a general-purpose modeling framework, and its societal impact depends on the context in which it is applied.

\section{Usage of Large Language Models}\label{app:sec:llms}
This work used large language models in the following ways:
\begin{description}
    \item[Preparation of plots] LLMs were partly used to generate the code for the plots presented in this work. The correctness of all code and data was checked manually. The data shown in the figures was generated by manually written code.
    \item[Code development] LLMs were used to assist in developing code for data generation, preparing procedural job launching scripts, documenting implemented functions or modules, and integrating existing GNN codebases into the working directory. All code modified by LLMs was manually checked. 
    \item[Polishing of manuscript] LLMs were occasionally used to refine or rephrase individual sentences.
\end{description}

\section{Graph Cut Penalty: Derivation and Gradient Analysis}
\label{app:cut}
Here we describe in more detail the graph cut penalty introduced in Section~\ref{sec:loss}.

\subsection{Loss definition}

For a matched pair with predicted membership probabilities
$p_u = \sigma(l_u)$ (where $\sigma$ is the sigmoid function and $l_u$ is the
per-node logit), the graph cut penalty is
\begin{equation}
  \mathcal{L}_\text{cut}
    = \frac{1}{|\mathcal{E}|}
      \sum_{(u,v)\in\mathcal{E}}
      \bigl[p_u(1-p_v)+p_v(1-p_u)\bigr].
  \label{eq:cut}
\end{equation}
Each summand is the product $p_u(1-p_v)$ (node $u$ inside, node $v$ outside)
plus $p_v(1-p_u)$ (the reverse), and equals zero when both endpoints are
on the same side of the boundary ($p \approx 0$ or $p \approx 1$) and
reaches its maximum of $\nicefrac{1}{2}$ when they straddle it
($p_u = p_v = \nicefrac{1}{2}$).

\subsection{Gradient analysis}

Differentiating~\eqref{eq:cut} with respect to logit $l_u$, and using
$\frac{\partial p_u}{\partial l_u} = p_u(1-p_u)$:
\begin{equation}
  \frac{\partial \mathcal{L}_\text{cut}}{\partial l_u}
  = \frac{p_u(1-p_u)}{|\mathcal{E}|}
    \sum_{v\colon(u,v)\in\mathcal{E}}
    (1 - 2p_v).
  \label{eq:cut_grad}
\end{equation}

Three structural properties of this gradient are noteworthy:
\begin{itemize}[leftmargin=1.5em, itemsep=2pt]
  \item \textbf{Saturated nodes are unaffected.}
    The pre-factor $p_u(1-p_u)$ is zero when $p_u \in \{0, 1\}$,
    so nodes that are already certain do not receive any gradient from this
    term, avoiding perturbation of confident predictions.

  \item \textbf{Majority-vote pressure.}
    The sum $\sum_v (1-2p_v)$ is negative when the majority of
    neighbors have $p_v > \nicefrac{1}{2}$ (i.e.\ are predicted
    inside the subgraph), and positive otherwise.
    A negative sum decreases the gradient, increasing $l_u$, hence pushing
    $u$ further inside; a positive sum pushes $u$ outward.
    The penalty therefore implements a differentiable
    neighbor-majority constraint.

  \item \textbf{Boundary sharpening.}
    The gradient is largest for boundary nodes ($p_u \approx \nicefrac{1}{2}$)
    that have disagreeing neighbors, and vanishes at topological interiors or exteriors, focusing learning signal exactly where structural ambiguity is highest.
\end{itemize}

\section{Dataset Descriptions}
\label{app:datasets}
We provide details of the creation of all datasets used in this work.

\subsection{Molecular datasets}

\begin{table}[thb]
    \centering
    \caption{\textbf{Statistics of the ChEMBL and ChEMBL12k dataset.}}\label{app:tab:chembl_stats}
    \resizebox{\textwidth}{!}{
    \begin{tabular}{lccccccc}
        \toprule
        Dataset & \# Graphs & \# Classes & Patterns/Graph & Nodes ($\mu\!\pm\!\sigma$) & Edges ($\mu\!\pm\!\sigma$) & Nodes [min,max] & Edges [min,max] \\
        \midrule
        ChEMBL & 2,827,875 & 95 & $5.8\pm 2.9$ & $29.8\pm 10.4$ & $32.4\pm 11.4$ & [1.0,100.0] & [0.0,132.0] \\
        ChEMBL12k & 12,000 & 95 & $5.8\pm 2.9$ & $29.9\pm 10.6$ & $32.5\pm 11.5$ & [5.0,100.0] & [4.0,108.0] \\
        \bottomrule
    \end{tabular}
    }
\end{table}

\paragraph{ChEMBL.}
We construct the ChEMBL dataset from ChEMBL-36~\citep{zdrazil2024chembl}
using a scalable preprocessing pipeline.
Starting from $2{,}848{,}825$ molecules, we retain those parseable by
RDKit and remove duplicates, yielding
$2{,}827{,}875$ molecules.
Functional group instances and their atom-level membership masks are
generated by Ertl's IFG algorithm~\citep{ertl2017ifg,colmenarejo2025efgs}, which provides
context-sensitive labeling (\eg distinguishing amide nitrogen from a free
amine).
After filtering functional group classes with frequency below $0.1\%$,
the vocabulary contains $95$ classes.
Molecules are featurized with OGB atom and bond features~\citep{hu2020ogb}
(9-dimensional atom features, 3-dimensional bond features).
The dataset is split 80/10/10 by Bemis--Murcko scaffold~\citep{bemis1996scaffold}
to ensure no training scaffold appears at test time.

\paragraph{ChEMBL12k.}
For ablation studies and rapid iteration, we define a 12{,}000-molecule
subset drawn uniformly at random from ChEMBL: $10{,}000$ molecules for
training, $1{,}000$ for validation, $1{,}000$ for test.
The subset shares the same atom/bond features and FG vocabulary as the full
ChEMBL dataset; the split is fixed by a global seed for reproducibility. The statistics of both datasets are provided in Table~\ref{app:tab:chembl_stats}.

\subsection{Synthetic datasets}

\begin{table}[thb]
\centering
\caption{\textbf{Statistics of all synthetic datasets.}}\label{app:tab:synthetic_stats}
\small
\resizebox{\textwidth}{!}{
\begin{tabular}{lccccccc}
\toprule
Dataset & \# Graphs & \# Classes & Patterns/Graph & Nodes ($\mu\!\pm\!\sigma$) & Edges ($\mu\!\pm\!\sigma$) & Nodes [min,max] & Edges [min,max] \\
\midrule
Cactus40      & 12000 & 6  & 6.3$\pm$1.2 & 40.0$\pm$0.0    & 45.3$\pm$1.2    & [40,40]     & [41,49]     \\
Cactus200     & 12000 & 6  & 4.8$\pm$0.9 & 200.0$\pm$0.0   & 203.8$\pm$0.9  & [200,200]   & [201,207]   \\
Cactus1000    & 12000 & 6  & 4.9$\pm$0.9 & 1000.0$\pm$0.0  & 1003.9$\pm$0.9  & [1000,1000] & [1001,1007] \\
Cliques100     & 12000 & 1 & 10.0$\pm$0.0 & 139.5$\pm$0.7   & 350.3$\pm$15.0 & [135,140]   & [298,412]   \\
Cliques200    & 12000 & 1  & 10.0$\pm$0.0 & 290.0$\pm$0.1   & 1450.3$\pm$30.0 & [288,290]   & [1346,1564] \\
Cliques400    & 12000 & 1  & 10.0$\pm$0.0 & 590.0$\pm$0.0   & 5900.2$\pm$60.2 & [590,590]   & [5643,6186] \\
ZINC12k       & 12000 & 9  & 5.5$\pm$3.3  & 23.2$\pm$4.5    & 24.9$\pm$5.3    & [9,37]      & [8,42]      \\
Mol-Reddit      & 11929 & 10 & 2.1$\pm$1.2 & 440.7$\pm$426.1 & 516.5$\pm$517.0  & [27,3810]   & [30,5333]   \\
Cactus fuzzy & 12000 & 4  & 4.4$\pm$1.3  & 100.0$\pm$0.0   & 104.7$\pm$0.9  & [100,100]   & [101,108]   \\
Cliques fuzzy& 12000 & 1  & 5.0$\pm$1.6  & 169.5$\pm$0.7   & 470.3$\pm$16.2   & [165,170]   & [404,534]   \\
\bottomrule
\end{tabular}
}
\end{table}

\begin{table}[ht]
  \centering
  \caption{%
    \textbf{Synthetic dataset query patterns.}
    \emph{Match}: whether instances are identified by exact subgraph
    isomorphism (exact) or approximate structural similarity (fuzzy).
    \emph{Node attr.}\ / \emph{Edge attr.}: feature type present on
    nodes / edges; \xmark{} = unlabeled (constant $\mathbf{1}$ feature).%
  }
  \label{app:tab:pattern_desc}
  \small
  \resizebox{\textwidth}{!}{%
  \begin{tabular}{lccccp{8cm}}
    \toprule
    \textbf{Dataset}
      & \textbf{\#Classes}
      & \textbf{Node attr.}
      & \textbf{Edge attr.}
      & \textbf{Match}
      & \textbf{Pattern description} \\
    \midrule
    Cactus40
      & 6
      & \xmark
      & \xmark
      & exact
      & Cycles of lengths 3--8 (one class per length), each embedded in a cactus host graph ($n{=}40$). \\
    Cactus200
      & 6
      & \xmark
      & \xmark
      & exact
      & Same as Cactus40 with cycle lengths 11--16
        and larger cactus hosts ($n{=}200$). \\
    Cactus1000
      & 6
      & \xmark
      & \xmark
      & exact
      & Same setup with cycle lengths 27--32
        within larger hosts ($n{=}1000$). \\
    \midrule
    Cliques100
      & 1
      & \xmark
      & \xmark
      & exact
      & Complete graph $K_5$ (5 nodes, 10 edges).
        Injected into random bipartite hosts ($n{=}50$ per part) by replacing a randomly chosen
        node with a 5-clique and rewiring its incident edges to uniformly
        random clique members. \\
    Cliques200
      & 1
      & \xmark
      & \xmark
      & exact
      & Complete graph $K_{10}$;
        same injection procedure, larger bipartite host ($n{=}100$ per part). \\
    Cliques400
      & 1
      & \xmark
      & \xmark
      & exact
      & Complete graph $K_{20}$ (20 nodes, 190 edges);
        same injection procedure, much larger bipartite host ($n{=}200$ per part). \\
    \midrule
    ZINC12k
      & 9
      & atom type
      & \xmark
      & exact
      & Nine frequent molecular subgraphs (3--7 nodes each) mined from
        ZINC12k by the GASTON algorithm~\citep{nijssen2004gaston} at minimum
        support 25\% and max occurrences of 5 per graph. Mined patterns are illustrated in Figure~\ref{app:fig:zinc12k}.
        Ground-truth instance masks are the GASTON-reported node sets for
        each occurrence within each molecule. \\ %
    Mol-Reddit
      & 10
      & \xmark
      & \xmark
      & exact
      & Ten molecular structures (19--62 nodes) with all atom and bond
        features discarded (pure topology) are randomly sampled from the Molhiv dataset of OGB~\citep{hu2020ogb}.
        Each structure is injected as a node-disjoint subgraph into a host
        graph drawn from REDDIT-MULTI-12K~\citep{yanardag2015deep}, with the number of attachment edges sampled from a Poisson distribution of mean 2.
        The host may contain unlabeled background motifs isomorphic to the
        injected patterns, requiring the model to distinguish true instances
        from background. Patterns are shown in Figure~\ref{app:fig:mol-reddit_patterns}. \\
    \midrule
    Cactus-fuzzy
      & 4
      & binary
      & \xmark
      & fuzzy
      & Target graphs are cactus graphs containing cycles of lengths $k \in \{7, \dots, 10\}$ as patterns. Each node receives a uniformly random label from $\{1,2\}$. The canonical pattern for length $k$ is a $k$-cycle with alternating labels $1$-$2$-$1$-$2$-$\cdots$. An cycle is annotated as an instance if its graph edit distance to the canonical pattern is at most $\tau_{\text{GED}}=3$; otherwise it is treated as background noise. \\
    Cliques-fuzzy
      & 1
      & \xmark
      & \xmark
      & fuzzy
      & The injection procedure follows the exact-match Cliques setting, but each injected $k$-clique is subsequently perturbed by randomly removing internal edges. Each injection is independently classified as an instance (with probability 50\%) or as noise. Instances have $4$ edges removed, while noise cliques have $8$ edges removed, yielding a more aggressive perturbation. Only instances are annotated with ground-truth masks; noise cliques are present in the target graphs but not annotated, requiring the model to distinguish between mild and heavy perturbations. \\
    \bottomrule
  \end{tabular}%
  }
\end{table}

All synthetic datasets are stored in Gaston format, with a target-graph file
and a pattern file.  Each sample is a host graph together with a set of
injected subgraph instances annotated by class label and node mask.
Statistics for each dataset are listed in Table~\ref{app:tab:synthetic_stats}. Query patterns for each dataset are described in Table~\ref{app:tab:pattern_desc}.

\paragraph{Cactus graphs.}
A cactus graph is generated by sampling a random labeled tree on $n$ nodes and injecting $c$ edge-disjoint cycles of lengths drawn uniformly from a specified range, by identifying tree paths of the required length and closing them into cycles. Each injected cycle forms a chordless induced subgraph by construction. The pattern vocabulary contains one class per distinct cycle length. Node and edge features are omitted (constant feature~$\mathbf{1}$).

\paragraph{Clique graphs.}
A host graph is a random bipartite graph $G(n_\text{side}, n_\text{side}, p)$
with $n_\text{side}$ nodes on each side.  Each clique injection replaces a
randomly chosen, previously unexpanded node $v$ with a $k$-clique, where
$v$'s neighbours are rewired to uniformly random nodes in the new clique,
and $v$ is deleted.  Disconnected graphs are resampled.
The single pattern class is the $k$-clique $K_k$.

\begin{figure}[htb]
    \centering
    \includegraphics[width=\linewidth]{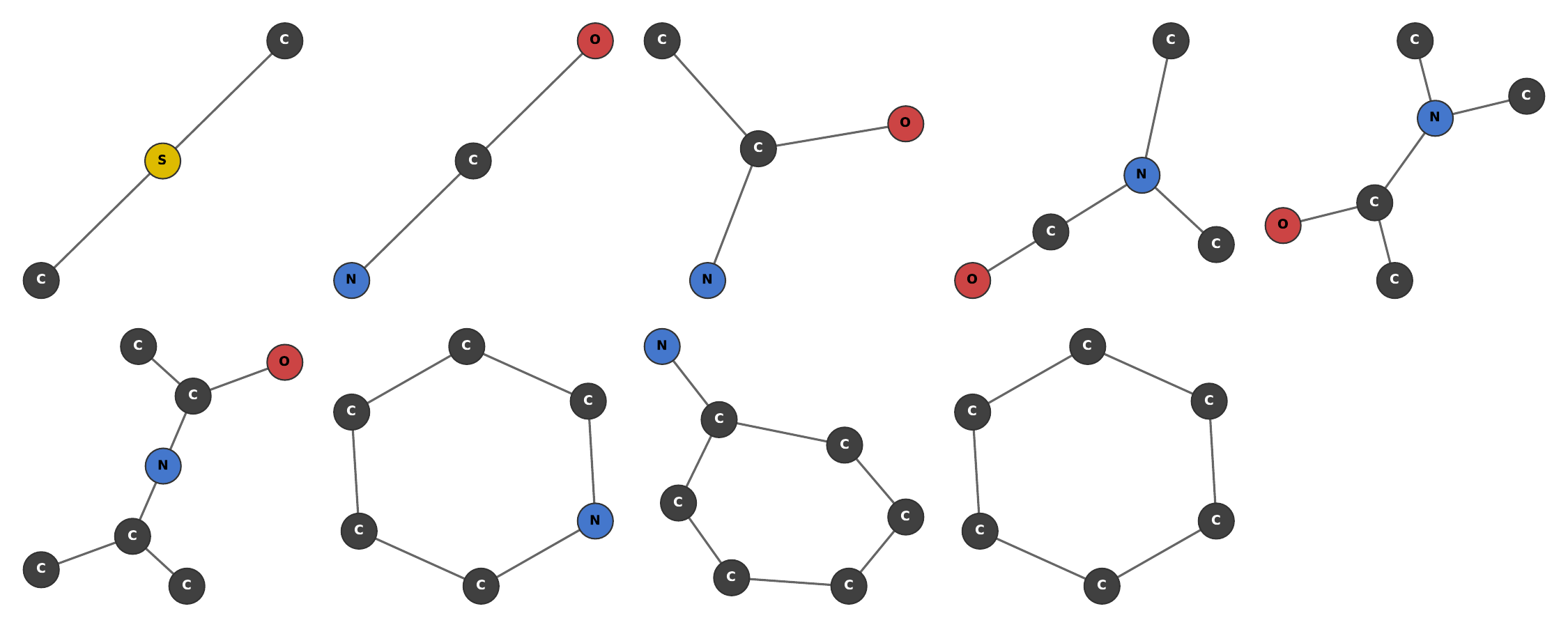}
    \caption{Query patterns for ZINC12k}
    \label{app:fig:zinc12k}
\end{figure}

\paragraph{ZINC12k.}
Frequent subgraph patterns are mined from the ZINC12k molecular benchmark using the GASTON algorithm~\citep{nijssen2004gaston} at minimum support 25\% and a maximum of 5 occurrences per graph. Nine high-frequency subgraphs with 3--7 atoms are retained as query patterns; only atom type is preserved as a node attribute. Ground-truth instance masks are the GASTON-reported node sets for each occurrence within each molecule. Mined patterns are visualized in Figure~\ref{app:fig:zinc12k}.

\begin{figure}[htb]
    \centering
    \includegraphics[width=\linewidth]{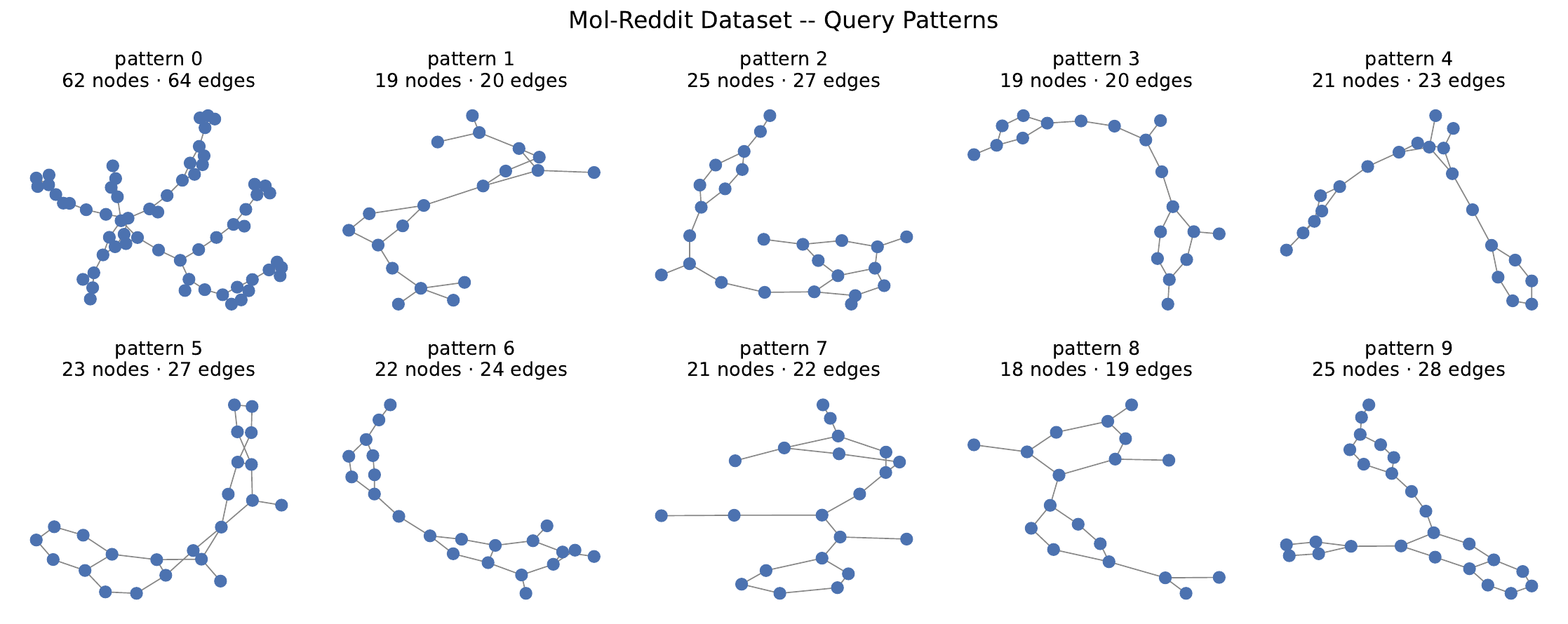}
    \caption{Query patterns for Mol-Reddit}
    \label{app:fig:mol-reddit_patterns}
\end{figure}

\paragraph{Mol-Reddit.}
Molecular structures (query patterns) are injected into social network host
graphs drawn from REDDIT-MULTI-12K~\citep{yanardag2015deep}.
Ten molecular structures are randomly sampled from the Molhiv dataset of OGB~\citep{hu2020ogb} as pattern classes;
their node features are discarded (constant~$\mathbf{1}$), leaving only
the topology for detection.
Each injection adds a molecular subgraph as a node-disjoint component of
the host graph, with $\sim$2\ (following a Poisson distribution) random attachment edges connecting the
molecule to the host.
This dataset emulates the scenario of incomplete annotations: the original
social network may contain graph motifs that are isomorphic to injected molecules,
which are not labeled, so the model must distinguish true injected instances
from background structure. The patterns are visualized in Figure~\ref{app:fig:mol-reddit_patterns}.

\paragraph{Fuzzy Cactus.}
Each node is assigned a binary label drawn uniformly from $\{1,2\}$.
The pattern class for a cycle of length $k$ is a \emph{canonical}
$k$-cycle with alternating labels $1$-$2$-$1$-$2$-$\cdots$.
An injected cycle is considered a \emph{proper} instance if its
node-label sequence differs from the canonical pattern (over all rotations
and reflections) by at most $\tau_\text{GED}$ substitutions;
otherwise it is treated as background noise.
Four cycle-length classes (7--10) are used, with the graph size set to
$n \approx 100$ nodes.

\paragraph{Fuzzy Cliques.}
The injection procedure follows the exact-match Cliques setting, but each
clique is subsequently perturbed by removing a random subset of its internal
edges, yielding an \emph{approximate} clique pattern.
A fraction of injections are designated as \emph{noise} (with probability
$p_\text{noise}=0.5$) and receive more aggressive edge removal; only proper
injections are annotated in the pattern file.
The single pattern class is a fuzzy $8$-clique ($k{=}8$) injected into
bipartite host graphs.

\section{Implementation Details}
\label{app:impl}

\subsection{Training setup}

All models are trained with the AdamW optimiser~\citep{loshchilov2018adamw}
with weight decay $10^{-4}$.
The learning rate follows a linear warm-up over 100 steps to the peak learning rate, followed by cosine decay.
Gradients are clipped to unit norm. All models are trained with PyTorch Lightning on a single GPU per run.

For the molecular task (ChEMBL12k), models are trained for up to 200 epochs
with batch size 128.
For synthetic tasks, epochs and batch sizes are adjusted per dataset owing
to the wide range of graph sizes. %
All results are reported as mean~$\pm$~std over 5 random seeds.

\subsection{Hyperparameters}\label{app:sec:hparams}

Table~\ref{app:tab:hparams_chembl} lists the hyperparameters used for the main
ChEMBL12k and ChEMBL experiments. Table~\ref{app:tab:hparams_synthetic} lists the search grid for the tuned hyperparameters; the other hyperparameters remain the same as for ChEMBL.

\begin{table}[htb]
  \centering
  \caption{Hyperparameters for the main \GraphDETR{} configuration on ChEMBL12k and ChEMBL.}
  \label{app:tab:hparams_chembl}
  \small
  \begin{tabular}{llp{8cm}}
    \toprule
    \textbf{Hyperparameter} & \textbf{Value} & \textbf{Description} \\
    \midrule
    \multicolumn{3}{l}{\textit{Model}} \\
    $D$ (hidden dim)              & 256  & Shared embedding dimension \\
    $Q$ (\# queries)             & 40   & Number of learnable query slots \\
    encoder blocks                & 5    & Number of graph encoder blocks for GCN/GIN/GraphGPS/NeuralWalker \\
    RWSE dim $K$                  & 20   & Random-walk structural encoding steps in the graph encoder \\
    \midrule
    \multicolumn{3}{l}{\textit{Two-way transformer decoder}} \\
    $L_\text{dec}$ (blocks)       & 3    & Number of two-way transformer decoder blocks \\
    $h$ (attention heads)         & 8    & Multi-head attention heads \\
    $d_\text{ff}$ (feedforward)   & 512  & MLP hidden dimension \\
    $r$ (downsample rate)         & 2    & Attention key/value dimension $= D/r$ \\
    \midrule
    \multicolumn{3}{l}{\textit{NeuralWalker encoder}} \\
    walk length $\ell$            & 20   & Walk sequence length \\
    walks per node (train)        & 1    & Random walks sampled per node \\
    walks per node (test)         & 5    & Walks at inference time \\
    sequence backend              & Mamba & SSM sequence model \\
    $d_\text{conv}$               & 9    & Conv1d kernel width \\
    $d_\text{state}$              & 16   & Mamba SSM state size \\
    window size $W$               & 8    & Structural encoding context \\
    bidirectional                 & \xmark & Forward-only walk processing \\
    \midrule
    \multicolumn{3}{l}{\textit{Loss}} \\
    $\lambda_\text{mask}$         & 4.0  & Mask loss total weight \\
    $\lambda_\text{cut}$          & 0.1  & Graph cut penalty relative weight \\
    $w_\varnothing$               & 0.1  & Background class CE weight \\
    \midrule
    \multicolumn{3}{l}{\textit{Optimisation}} \\
    optimiser                     & AdamW   & \\
    learning rate                 & $10^{-3}$ & Peak learning rate \\
    weight decay                  & $10^{-4}$ & AdamW weight decay \\
    warmup steps                  & 100       & Linear warm-up steps \\
    max epochs                    & 200       & ChEMBL12k training budget \\
    gradient clip norm            & 1.0       & \\
    batch size                    & 128       & Graphs per batch \\
    \bottomrule
  \end{tabular}
\end{table}

\begin{table}[htb]
    \centering
    \caption{Hyperparameter search grids for \GraphDETR{} on synthetic datasets.}
    \label{app:tab:hparams_synthetic}
    \begin{tabular}{lp{8cm}}\toprule
        \textbf{Hyperparameter} & \textbf{Search grid or values} \\ \midrule
        $Q$ (\# queries) & $\{20, 40, 60 \}$ \\
        $\lambda_\text{cut}$ & $\{0.0, 0.001, 0.01, 0.1\}$\\
        batch size & we adjust it based on the graph sizes in the dataset \\
        NeuralWalker walk length & 20 for Cactus40, ZINC12k, Cactus fuzzy, 5 for Cliques100, Cliques200, Cliques400, Cliques fuzzy, 50 for Mol-Reddit, 200 for Cactus200, 1000 for Cactus1000. \\
        \bottomrule
    \end{tabular}
\end{table}

\subsection{Computing resources}\label{app:sec:computing}

All experiments are run on a GPU cluster with NVIDIA A100 or H100 GPUs.
A single ChEMBL12k run (200 epochs) completes in under 8 hours on one GPU.
A full ChEMBL run (5 encoder layers, batch size 128) takes approximately
one day on a single GPU.
Synthetic dataset runs vary from a few hours (Cactus40, Cliques100)
to over one day (Cactus1000, Mol-Reddit) depending on graph size.

\section{Additional Results}
\label{app:results}

\subsection{Full synthetic results}

Table~\ref{tab:synthetic_full} reports all five evaluation metrics for every
encoder--dataset combination.

\begin{table}[ht]
  \centering
  \caption{%
    \textbf{Full synthetic results.}
    All five metrics for every encoder--dataset combination
    (mean\,$\pm$\,std over 5 seeds).%
  }
  \label{tab:synthetic_full}
  \small
  \resizebox{.95\textwidth}{!}{%
  \begin{tabular}{llccccc}
    \toprule
    \textbf{Dataset} & \textbf{Encoder}
      & \textbf{AP$_{100}$} & \textbf{mAP}
      & \textbf{Rec$_{100}$} & \textbf{mIoU} & \textbf{ExactMatch} \\
    \midrule
    \multirow{4}{*}{Cactus40}
      & GCN          & $92.33_{\pm 0.56}$ & $95.55_{\pm 0.19}$ & $95.37_{\pm 0.26}$ & $99.14_{\pm 0.05}$ & $81.83_{\pm 2.30}$ \\
      & GIN          & $92.98_{\pm 0.90}$ & $95.98_{\pm 0.42}$ & $95.76_{\pm 0.42}$ & $99.21_{\pm 0.11}$ & $83.77_{\pm 1.32}$ \\
      & GraphGPS     & $96.51_{\pm 0.43}$ & $97.84_{\pm 0.21}$ & $97.82_{\pm 0.23}$ & $99.57_{\pm 0.04}$ & $89.63_{\pm 0.83}$ \\
      & NeuralWalker & $97.44_{\pm 0.44}$ & $98.27_{\pm 0.24}$ & $98.66_{\pm 0.29}$ & $99.77_{\pm 0.04}$ & $84.02_{\pm 1.79}$ \\
    \midrule
    \multirow{4}{*}{Cactus200}
      & GCN          & $42.80_{\pm 3.16}$ & $57.86_{\pm 5.04}$ & $55.65_{\pm 3.87}$ & $93.70_{\pm 0.94}$ & $18.12_{\pm 10.59}$ \\
      & GIN          & $41.51_{\pm 4.45}$ & $52.94_{\pm 3.82}$ & $53.47_{\pm 3.04}$ & $93.84_{\pm 2.06}$ & $9.20_{\pm 4.25}$ \\
      & GraphGPS     & $83.67_{\pm 0.78}$ & $92.17_{\pm 0.64}$ & $90.44_{\pm 0.55}$ & $98.96_{\pm 0.06}$ & $84.12_{\pm 1.16}$ \\
      & NeuralWalker & $91.90_{\pm 2.41}$ & $95.78_{\pm 1.60}$ & $94.08_{\pm 1.55}$ & $99.34_{\pm 0.17}$ & $71.98_{\pm 13.66}$ \\
    \midrule
    \multirow{4}{*}{Cactus1000}
      & GCN          & $10.46_{\pm 3.04}$ & $17.75_{\pm 4.38}$ & $18.44_{\pm 3.01}$ & $89.25_{\pm 1.46}$ & $0.00_{\pm 0.00}$ \\
      & GIN          & $4.82_{\pm 7.30}$  & $7.83_{\pm 11.85}$ & $7.97_{\pm 11.74}$ & $35.37_{\pm 48.44}$ & $0.07_{\pm 0.15}$ \\
      & GraphGPS     & $40.22_{\pm 0.68}$ & $47.00_{\pm 0.26}$ & $45.69_{\pm 0.29}$ & $94.86_{\pm 0.31}$ & $3.70_{\pm 0.40}$ \\
      & NeuralWalker & $42.20_{\pm 2.06}$ & $61.51_{\pm 2.58}$ & $57.81_{\pm 2.08}$ & $94.06_{\pm 0.44}$ & $0.32_{\pm 0.37}$ \\
    \midrule
    \multirow{4}{*}{Cliques100}
      & GCN          & $90.94_{\pm 2.48}$ & $94.03_{\pm 1.56}$ & $93.53_{\pm 1.75}$ & $98.73_{\pm 0.34}$ & $69.90_{\pm 7.65}$ \\
      & GIN          & $93.31_{\pm 0.74}$ & $96.03_{\pm 0.47}$ & $95.52_{\pm 0.57}$ & $99.04_{\pm 0.11}$ & $88.23_{\pm 4.25}$ \\
      & GraphGPS     & $97.05_{\pm 1.60}$ & $98.16_{\pm 0.79}$ & $98.02_{\pm 1.17}$ & $99.58_{\pm 0.25}$ & $96.00_{\pm 4.74}$ \\
      & NeuralWalker & $96.15_{\pm 0.57}$ & $97.40_{\pm 0.41}$ & $97.71_{\pm 0.39}$ & $99.49_{\pm 0.09}$ & $95.93_{\pm 1.06}$ \\
    \midrule
    \multirow{4}{*}{Cliques200}
      & GCN          & $90.33_{\pm 1.46}$ & $95.21_{\pm 0.90}$ & $93.91_{\pm 0.59}$ & $99.11_{\pm 0.12}$ & $86.47_{\pm 10.14}$ \\
      & GIN          & $92.84_{\pm 2.20}$ & $96.69_{\pm 1.04}$ & $95.42_{\pm 1.85}$ & $99.31_{\pm 0.27}$ & $95.63_{\pm 2.39}$ \\
      & GraphGPS     & $96.45_{\pm 1.17}$ & $98.23_{\pm 0.50}$ & $97.94_{\pm 0.88}$ & $99.72_{\pm 0.14}$ & $97.92_{\pm 1.64}$ \\
      & NeuralWalker & $93.48_{\pm 2.35}$ & $97.17_{\pm 0.77}$ & $96.08_{\pm 1.41}$ & $99.45_{\pm 0.17}$ & $97.18_{\pm 1.92}$ \\
    \midrule
    \multirow{4}{*}{Cliques400}
      & GCN          & $92.51_{\pm 2.85}$ & $95.91_{\pm 2.54}$ & $95.11_{\pm 2.07}$ & $99.37_{\pm 0.33}$ & $87.28_{\pm 15.53}$ \\
      & GIN          & $94.12_{\pm 1.51}$ & $97.49_{\pm 0.54}$ & $96.31_{\pm 0.58}$ & $99.63_{\pm 0.06}$ & $95.42_{\pm 1.13}$ \\
      & GraphGPS     & $95.05_{\pm 3.23}$ & $98.14_{\pm 0.99}$ & $97.06_{\pm 2.34}$ & $99.72_{\pm 0.28}$ & $98.35_{\pm 2.25}$ \\
      & NeuralWalker & $92.58_{\pm 3.48}$ & $97.53_{\pm 0.84}$ & $95.77_{\pm 1.66}$ & $99.60_{\pm 0.16}$ & $97.78_{\pm 0.66}$ \\
    \midrule
    \multirow{4}{*}{ZINC12k}
      & GCN          & $89.68_{\pm 0.88}$ & $93.78_{\pm 0.47}$ & $93.47_{\pm 0.52}$ & $98.71_{\pm 0.14}$ & $84.70_{\pm 2.34}$ \\
      & GIN          & $92.86_{\pm 0.25}$ & $95.60_{\pm 0.10}$ & $95.33_{\pm 0.15}$ & $99.11_{\pm 0.04}$ & $88.73_{\pm 1.33}$ \\
      & GraphGPS     & $95.28_{\pm 0.13}$ & $97.15_{\pm 0.07}$ & $97.08_{\pm 0.24}$ & $99.38_{\pm 0.08}$ & $92.58_{\pm 2.64}$ \\
      & NeuralWalker & $97.66_{\pm 0.27}$ & $98.60_{\pm 0.16}$ & $98.79_{\pm 0.18}$ & $99.75_{\pm 0.04}$ & $94.73_{\pm 1.33}$ \\
    \midrule
    \multirow{4}{*}{Mol-Reddit}
      & GCN          & $65.86_{\pm 1.11}$ & $85.05_{\pm 0.67}$ & $76.04_{\pm 0.59}$ & $96.51_{\pm 0.20}$ & $93.17_{\pm 0.83}$ \\
      & GIN          & $62.12_{\pm 1.31}$ & $82.67_{\pm 0.92}$ & $73.45_{\pm 0.93}$ & $96.15_{\pm 0.24}$ & $91.89_{\pm 0.82}$ \\
      & GraphGPS     & $80.78_{\pm 1.32}$ & $90.67_{\pm 1.08}$ & $84.59_{\pm 1.44}$ & $97.51_{\pm 0.28}$ & $95.26_{\pm 1.12}$ \\
      & NeuralWalker & $89.76_{\pm 0.97}$ & $95.26_{\pm 0.42}$ & $93.12_{\pm 0.60}$ & $98.99_{\pm 0.13}$ & $97.30_{\pm 0.16}$ \\
    \midrule
    \multirow{4}{*}{Cactus-fuzzy}
      & GCN          & $67.78_{\pm 1.17}$ & $78.07_{\pm 1.08}$ & $85.55_{\pm 0.92}$ & $98.15_{\pm 0.07}$ & $35.83_{\pm 2.91}$ \\
      & GIN          & $76.19_{\pm 2.03}$ & $84.19_{\pm 2.42}$ & $89.56_{\pm 0.78}$ & $98.70_{\pm 0.10}$ & $45.15_{\pm 5.97}$ \\
      & GraphGPS     & $84.86_{\pm 0.51}$ & $91.34_{\pm 0.29}$ & $93.57_{\pm 0.39}$ & $99.15_{\pm 0.16}$ & $79.95_{\pm 1.18}$ \\
      & NeuralWalker & $90.91_{\pm 2.88}$ & $93.19_{\pm 2.92}$ & $97.78_{\pm 0.35}$ & $99.77_{\pm 0.04}$ & $43.38_{\pm 15.89}$ \\
    \midrule
    \multirow{4}{*}{Cliques-fuzzy}
      & GCN          & $95.05_{\pm 1.02}$ & $96.81_{\pm 0.57}$ & $96.83_{\pm 0.45}$ & $99.45_{\pm 0.06}$ & $90.82_{\pm 1.37}$ \\
      & GIN          & $95.26_{\pm 0.89}$ & $97.22_{\pm 0.46}$ & $96.48_{\pm 0.62}$ & $99.38_{\pm 0.15}$ & $97.37_{\pm 0.85}$ \\
      & GraphGPS     & $96.66_{\pm 0.86}$ & $98.05_{\pm 0.48}$ & $98.13_{\pm 0.68}$ & $99.65_{\pm 0.13}$ & $99.15_{\pm 0.55}$ \\
      & NeuralWalker & $93.94_{\pm 2.10}$ & $96.58_{\pm 0.94}$ & $96.32_{\pm 1.28}$ & $99.35_{\pm 0.20}$ & $96.30_{\pm 1.91}$ \\
    \bottomrule
  \end{tabular}%
  }
\end{table}

\subsection{Qualitative results on ChEMBL}
\label{app:sec:qualitative}
\begin{figure}[htb]
    \centering
    \includegraphics[width=0.7\linewidth]{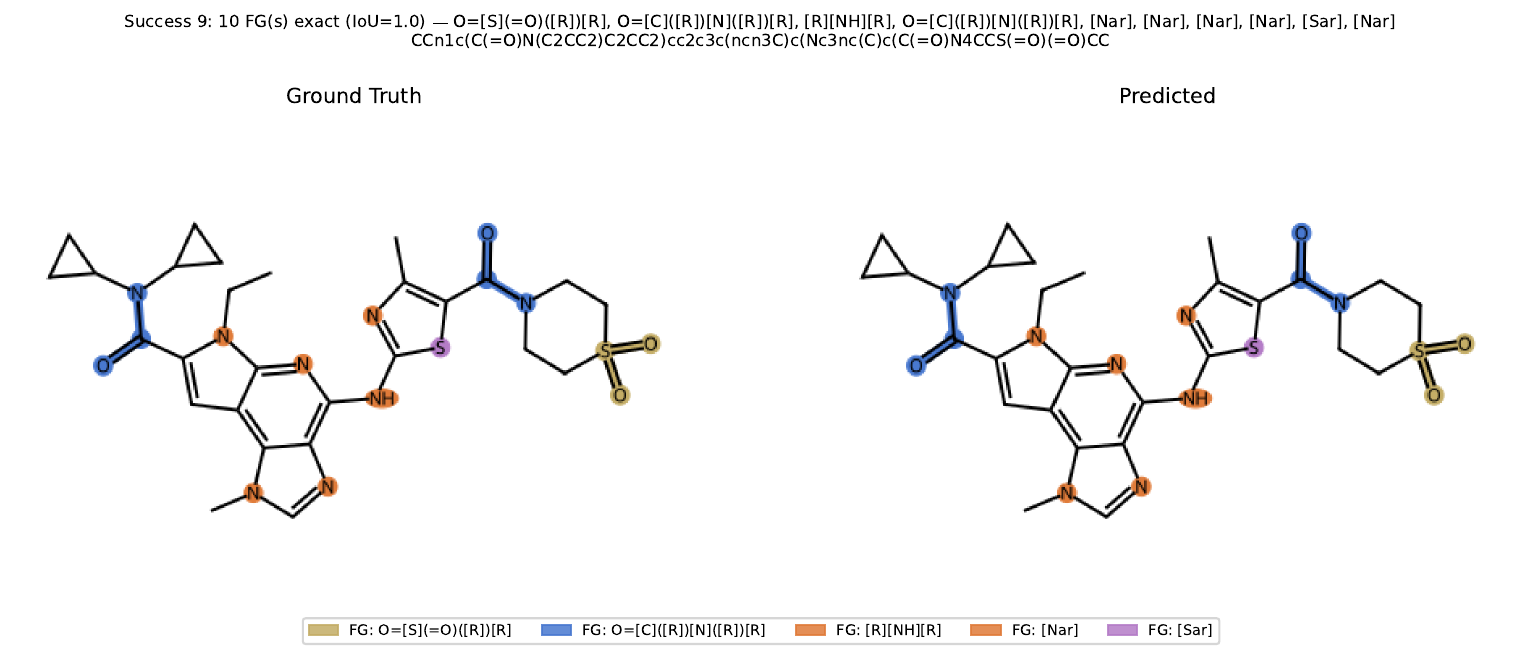}
    \includegraphics[width=0.7\linewidth]{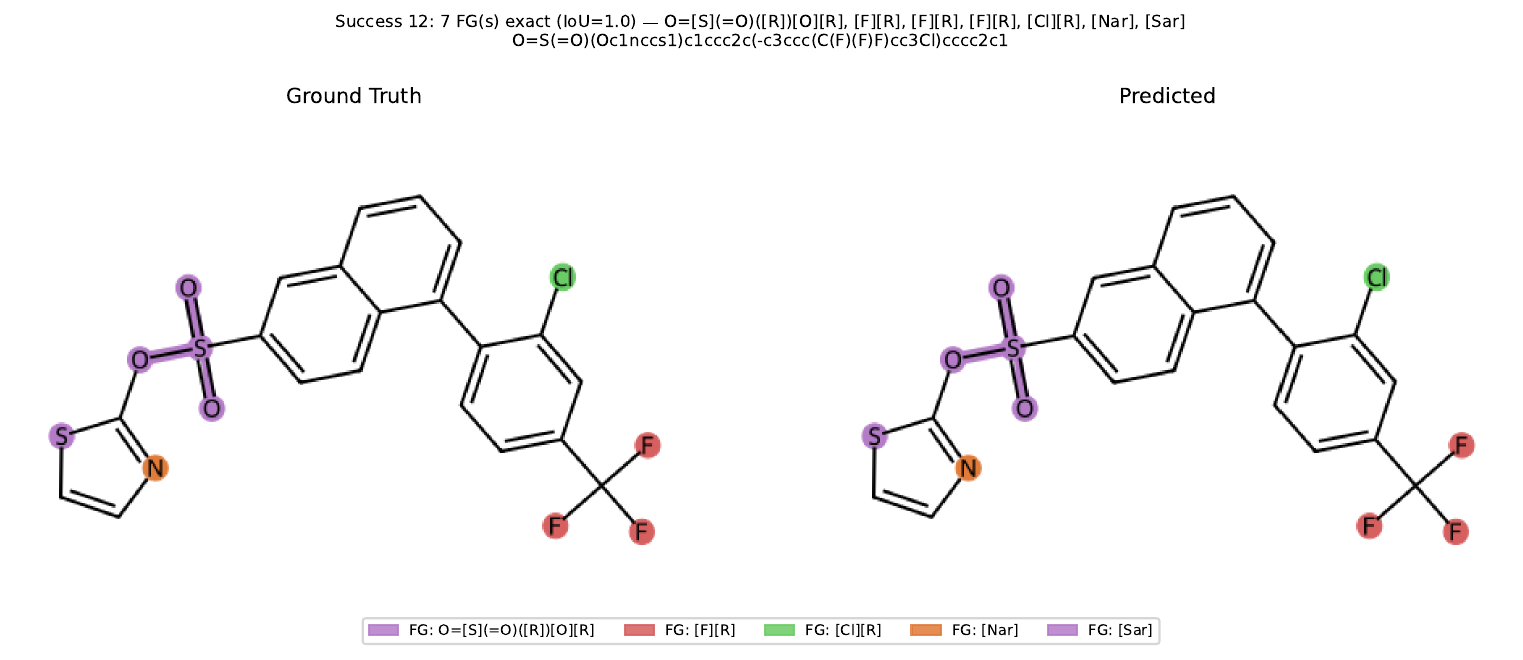}
    \includegraphics[width=0.7\linewidth]{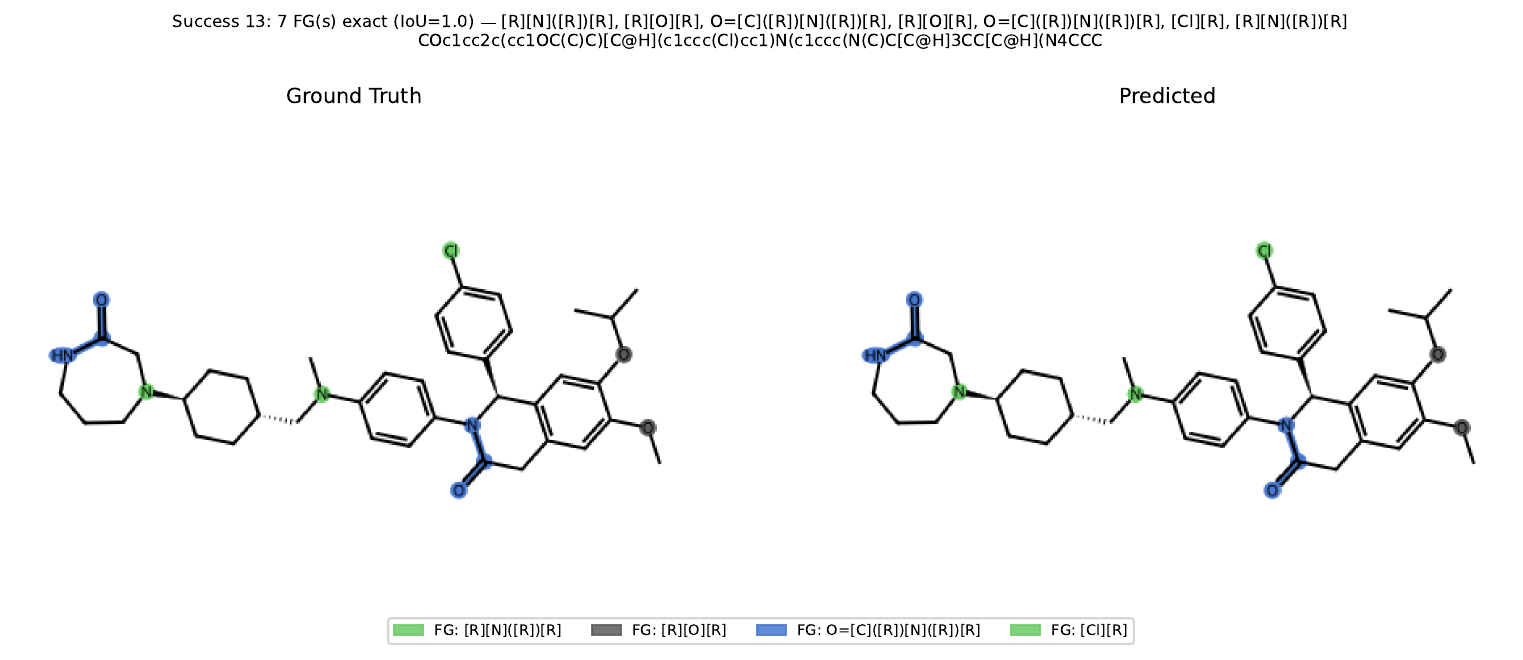}
    \includegraphics[width=0.7\linewidth]{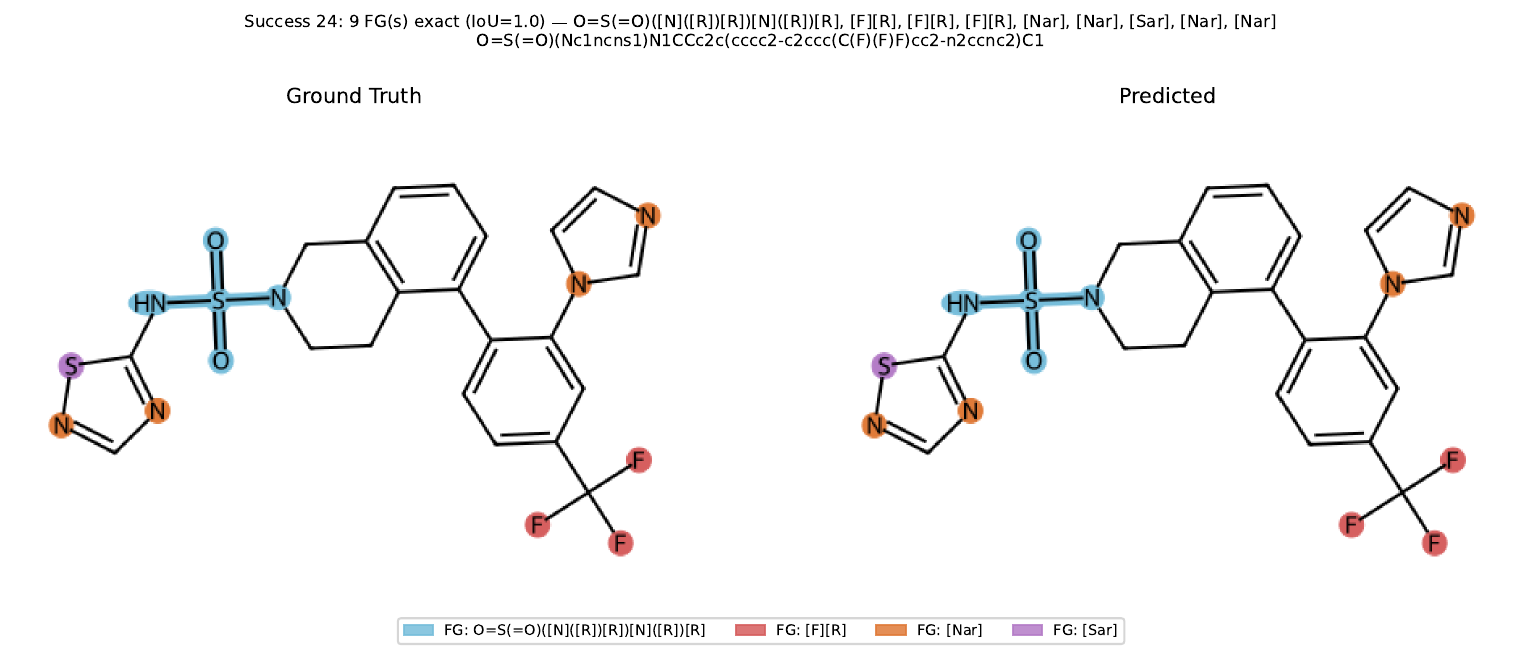}
    \includegraphics[width=0.7\linewidth]{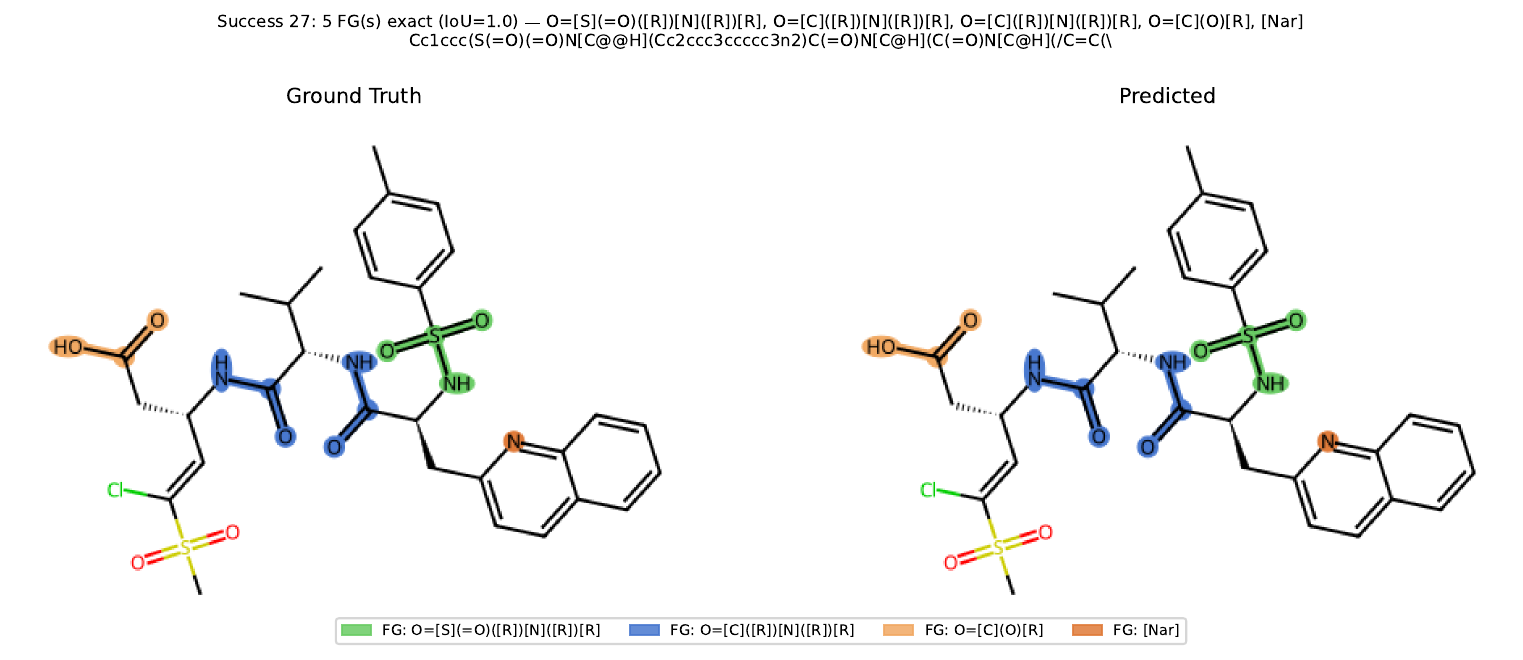}
    \caption{Visualization of ChEMBL examples whose functional groups are successfully predicted by \GraphDETR{}. }
    \label{app:fig:example_chembl}
\end{figure}

We provide in Figure~\ref{app:fig:example_chembl} a few examples from the ChEMBL dataset whose functional groups are successfully predicted by \GraphDETR{}.

\end{document}